\documentclass[10pt,twocolumn,letterpaper]{article}

\usepackage{cvpr}
\usepackage{times}
\usepackage{epsfig}
\usepackage{graphicx}
\usepackage{amsmath}
\usepackage{amssymb}

\usepackage[numbers,sort]{natbib}
\usepackage{xcolor, colortbl}
\usepackage{tabularx}
\usepackage{multirow}

\newcolumntype{C}[1]{>{\centering\let\newline\\\arraybackslash\hspace{0pt}}p{#1}}

\def\ie{\emph{i.e.}}
\def\eg{\emph{e.g.}}
\def\etal{\emph{et al.}}

\definecolor{brown}{rgb}{0.65, 0.16, 0.16}
\definecolor{purp}{rgb}{0.65, 0.16, 0.65}
\newcommand{\suha}[1]{{\color{brown}{[#1]}}}
\newcommand{\jw}[1]{{\color{purp}{[#1]}}}

\usepackage[pagebackref=true,breaklinks=true,letterpaper=true,colorlinks,bookmarks=false]{hyperref}

\cvprfinalcopy % *** Uncomment this line for the final submission

\def\cvprPaperID{420} % *** Enter the CVPR Paper ID here

\ifcvprfinal\fi
\begin{document}

%%%%%%%%% TITLE
\title{Learning Pixel-level Semantic Affinity with Image-level Supervision\\for Weakly Supervised Semantic Segmentation}
%reciprocal,coupled,part-to-part,lumped,region-basd,part-based,
%agglomerated,clustered,grafted,glued,
%interosculated,inosculated,osculated

% learning semi-pixel-level semantic affinity with image-level supervision for weakly supervised semantic segmentation

\author{Jiwoon Ahn\\
DGIST, Korea\\
{\tt\small jyun@dgist.ac.kr}
\and
Suha Kwak\\
POSTECH, Korea\\
{\tt\small suha.kwak@postech.ac.kr}
}

\maketitle

\pagestyle{plain}

%%% ABSTRACT %%%%%%%%%%%%%%%%%%%%%%%%%%%%%%%%%%%%%%%%%%%%
% !TEX root = weaksup_jiwoon.tex

\begin{abstract}

The deficiency of segmentation labels is one of the main obstacles to semantic segmentation in the wild.
To alleviate this issue, we present a novel framework that generates segmentation labels of images given their image-level class labels.
In this weakly supervised setting, trained models have been known to segment local discriminative parts rather than the entire object area.
Our solution is to propagate such local responses to nearby areas which belong to the same semantic entity.
To this end, we propose a Deep Neural Network (DNN) called AffinityNet that predicts semantic affinity between a pair of adjacent image coordinates. 
The semantic propagation is then realized by random walk with the affinities predicted by AffinityNet.
More importantly, the supervision employed to train AffinityNet is given by the initial discriminative part segmentation, which is incomplete as a segmentation annotation but sufficient for learning semantic affinities within small image areas.
Thus the entire framework relies only on image-level class labels and does not require any extra data or annotations.
On the PASCAL VOC 2012 dataset, a DNN learned with segmentation labels generated by our method outperforms previous models trained with the same level of supervision, and is even as competitive as those relying on stronger supervision.

\end{abstract}

%%% INTRODUCTION %%%%%%%%%%%%%%%%%%%%%%%%%%%%%%%%%%%%%%%%
% !TEX root = weaksup_jiwoon.tex

\section{Introduction}
\label{sec:intro}

% ======================================================================
% \recap{semantic segmentation and lack of annotated data}
% - recent progress with DNNs
% - "lack of annotated examples" issue
% ======================================================================
Recent development of Deep Neural Networks (DNNs) has driven the remarkable improvements in semantic segmentation~\cite{Fcn,Deeplabcrf,Lin16,Qi2016,Crfrnn,deconvnet,Bertasius_2017_CVPR,ChenTPAMI17}.
Despite the great success of DNNs, however, we still have a far way to go in achieving semantic segmentation in an uncontrolled and realistic environment.
% One of the main obstacles is the lack of annotated training data, caused by the prohibitively expensive annotation cost of pixel-level segmentation labels.
% For this reason, existing datasets often suffer from lack of annotated examples and class diversity, and it is not straightforward to learn DCNNs for semantic segmentation of objects out of the classes in the datasets.
One of the main obstacles is \emph{lack of training data}.
Due to the prohibitively expensive annotation cost of pixel-level segmentation labels, existing datasets often suffer from lack of annotated examples and class diversity.
% Since the conventional DNN-based approaches demand the pixel-level labels as supervision, it is not straightforward to them to be extended to handle more classes 
% For this reason, in the conventional approaches, it is not straightforward to learn DNNs for semantic segmentation of objects out of the predefined classes in the datasets.
This makes the conventional approaches limited to a small range of object categories predefined in the datasets. %, and bounds their performance as well. 

% ======================================================================
% \recap{weakly supervised semantic segmentation}
% - a solution to the above issue
% - prev approaches - stronger than image-level class labels
%   -- bounding box
%   -- scribble
%   -- point
%   -- limitations:
%       = they are typically not readily available
%       = sometimes require additional human intervention
% - prev approaches - image-level class labels
%   -- the minimum level of supervision we can utilized
%   -- readily available in large scale image datasets and/or the Web
%   -- low-level image structure~\cite{KwakAAAI17}
%   -- motions in video~\cite{HongCVPR17}
%   -- saliency map~\cite{OhCVPR17}
%   -- limitations:
%       = require preprocessing like superpixel, motion, and saliency
%       = the above stuffs are heavy, off-the-shelf, and not trainable
% ======================================================================
Weakly supervised approaches have been studied to resolve the above issue and allow semantic segmentation models more scalable. 
% Their common motivation is to make use of an enormous amount of visual data for which weaker annotations like bounding boxes~\cite{Boxsup,wssl,KhorevaCVPR17} and scribbles~\cite{scribblesup} are readily available.
Their common motivation is to utilize annotations like bounding boxes~\cite{Boxsup,wssl,KhorevaCVPR17} and scribbles~\cite{scribblesup,Vernaza_2017_CVPR} that are weaker than pixel-level labels but readily available in a large amount of visual data or easily obtainable thanks to their low annotation costs.
Among various types of weak annotations for semantic segmentation, image-level class labels have been widely used~\cite{KwakAAAI17,HongCVPR17,Wsl,OhCVPR17,Ccnn,sec,WeiCVPR17} since they are already given in existing large-scale image datasets (\eg, ImageNet~\cite{Imagenet}) or automatically annotated for image retrieval results by search keywords.
% and automatically annotated to web image search results by search keywords.
However, learning semantic segmentation with the image-level label supervision is a significantly ill-posed problem since such supervision indicates only the existence of a certain object class, and does not inform object location and shape that are essential for learning segmentation.
% Indeed, the key determinant of success in this line of research is how well the missing information is compensated.

% to this end, approaches in this line of research have synthesized 

Approaches in this line of research have incorporated additional evidences to simulate the location and shape information absent in the supervision.
A popular choice for the localization cue is the Class Activation Map (CAM)~\cite{Cam}, which highlights local discriminative parts of target object by investigating the contribution of hidden units to the output of a classification DNN.
The discriminative areas highlighted by CAMs are in turn used as seeds that will be propagated to cover the entire object area.
To recover the object area accurately from the seeds, previous approaches have utilized image segmentation~\cite{KwakAAAI17,Wsl}, motions in video~\cite{Tokmakov16}, or both~\cite{HongCVPR17}, all of which are useful to estimate object shape.
For the same purpose, a class-agnostic salient region is estimated and incorporated with the seeds in~\cite{OhCVPR17}.
However, they demand extra data (\ie, videos)~\cite{HongCVPR17,Tokmakov16}, additional supervision (\ie, object bounding box)~\cite{OhCVPR17}, or off-the-shelf techniques (\ie, image segmentation) that cannot take advantage of representation learning in DNNs~\cite{HongCVPR17,KwakAAAI17,Wsl}.

%% ======================================================================
%% FIGURE START
\begin{figure*} [!t]
\centering
\includegraphics[width = 0.95 \textwidth]{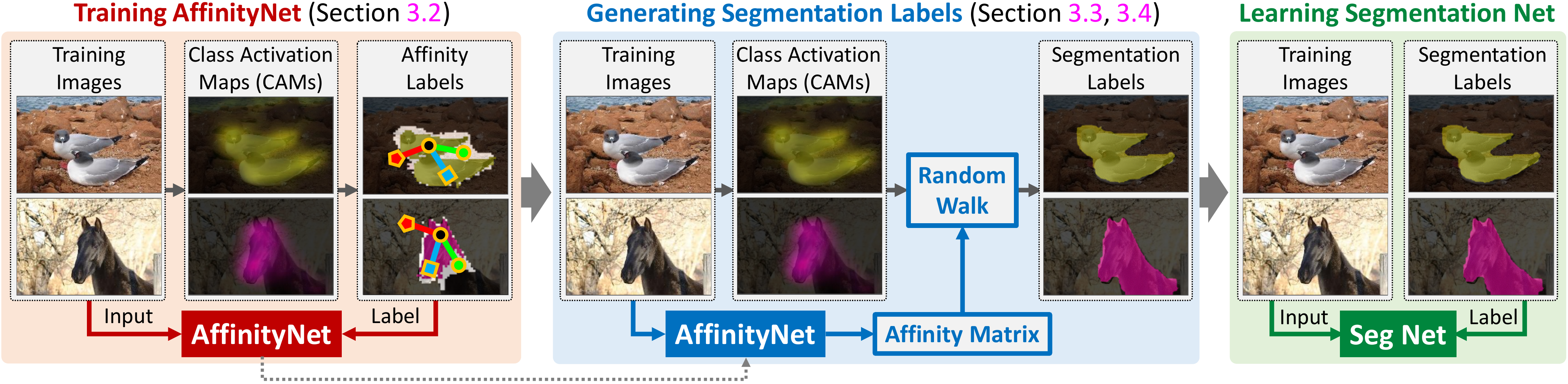}
\caption{Illustration of our approach.
Salient areas for object classes and background are first localized in training images by CAMs~\cite{Cam} (Section~\ref{sec:cam}).
From the salient regions, we sample pairs of adjacent coordinates
and assign binary labels to them according to their class consistency.
% for which it is clear whether their classes are the same or not,
% which are likely to be semantically equivalent or different, 
The labeled pairs are then used to train AffinityNet (Section~\ref{sec:learn_affinitynet}).
The trained AffinityNet in turn predicts semantic affinities within local image areas, which are incorporated with random walk to revise the CAMs (Section~\ref{sec:revising_cams}) and generate their segmentation labels (Section~\ref{sec:learn_segnet}).
Finally, the generated annotations are employed as supervision to train a semantic segmentation model.
} \vspace{-0.2cm}
\label{fig:outline}
\end{figure*}
%% FIGURE END
%% ======================================================================

% ======================================================================
% \recap{our approach}
% - simple yet effective weaksup (image label only) approach based on DCNNs
% - overall framework, in a nutshell
%   1) estimate initial seed of segmentation through CAM~\cite{Cam}
%   2) compute semantic affinity between two adjacent positions in image
%   3) propagate the seeds via Random Walk with the estimated affinity 
% - key idea:
%   -- learning the semantic affinity from incomplete annotation (CAM)
% - contributions:
%   -- not rely on off-the-shelf technique
%   -- learning affinity from incomplete annotation with a DCNN
%   -- state of the art in weakly supervised semantic segmentation
% ======================================================================
% - connect each pixel with local neighbors 
% - predict 
% - learning affinity
% - random walk
% - fill the gap, and suppress false positives
In this paper, we present a simple yet effective approach to compensate the missing information for object shape with no external data or additional supervision.
The key component of our framework is AffinityNet, which is a DNN that takes an image as input and predicts semantic affinities of pairs of adjacent image coordinates.
Given an image and its CAMs, we first build a neighborhood graph where each pixel is connected to its neighbors within a certain radius, and estimate semantic affinities of pairs connected in the graph through AffinityNet.
Sparse activations in CAMs are then diffused by random walk~\cite{randomwalk} on the graph, for each class:
The affinities on edges in the graph encourage random walk to propagate the activations to nearby and semantically identical areas, and penalize propagation to areas of the other classes.
This semantic diffusion revises CAMs significantly so that fine object shapes are recovered. 
% We apply this process to training images for generating segmentation labels from their revised CAMs, and the segmentation labels are used to train a segmentation model.
% We apply this process to training images for computing their revised CAMs and synthesizing segmentation labels by taking the class label associated to the maximum activation on the CAMs at each pixel.
We apply this process to training images for synthesizing their segmentation labels by taking the class label associated to the maximum activation of the revised CAMs at each pixel.
The generated segmentation labels are used to train a segmentation model for testing.

% - getting hints from segmentation seeds of the other images.
% - segmentation seeds are obviously not sufficiently accurate as supervision for semantic segmentation, but good enough to supervise AffinityNet (affinities within a small image area)
The remaining issue is how to learn AffinityNet without extra data or additional supervision.
% We utilize the initial CAMs of training images as sources of supervision.
To this end, the initial CAMs of training images are utilized as sources of supervision.
Because CAMs often miss some object parts and exhibit false alarms,
they are incomplete as a supervision for learning semantic segmentation whose goal is to predict the entire object masks accurately.
However, we found that they are often locally correct and provide evidences to identify semantic affinities within a small image area, which are the objective of AffinityNet.
% However, we found that they are sufficient to supervise the local semantic affinities within a small image area, which are the objective of AffinityNet, since they are often locally correct.
%To generate reliable labels of the local semantic affinities, we first remove activation scores irrelevant to the groundtruth image-level labels from CAMs. - \jw{This step is also applied to the seeds for propagation. So may confuse readers.}
%Furthermore, we disregard uncertain areas of CAMs by conservative thresholding so that only confident object and background areas are segmented. %\suha{it is not actually a simple thresholding... must be re-written}
% \jw{Since this procedure is nothing but simple, I think we should only briefly illustrate.}
To generate reliable labels of the local semantic affinities, 
% \jw{we disregard uncertain areas of CAMs by sorting out regions with relatively high scores of CAMs} so that only confident object and background areas are segmented.
we disregard areas with relatively low activation scores on the CAMs so that only confident object and background areas remain.
A training example is then obtained by sampling a pair of adjacent image coordinates on the confident areas, and its binary label is 1 if its coordinates belong to the same class and 0 otherwise.

The overall pipeline of the proposed approach is illustrated in Figure~\ref{fig:outline}.
First, CAMs of training images are computed and utilized to generate semantic affinity labels, which are used as supervision to train AffinityNet.
We then apply the trained AffinityNet to each training image to compute the semantic affinity matrix of its neighborhood graph, which is employed in random walk to revise its CAMs and obtain synthesized segmentation labels.
Finally, the generated segmentation labels are used to train a semantic segmentation DNN, which is the only network that will be used at test time.
Our contribution is three-fold:
\begin{itemize}
% \item We propose a novel DNN named AffinityNet that estimates semantic affinities of pairs of adjacent image coordinates. Furthermore, it requires only image-level label supervision to be trained.
\item We propose a novel DNN named AffinityNet that predicts high-level semantic affinities in a pixel-level, but is trained with image-level class labels only.
\item Unlike most previous weakly supervised methods, our approach does not rely heavily on off-the-shelf techniques, and takes advantage of representation learning through end-to-end training of AffinityNet.
\item On the PASCAL VOC 2012~\cite{Pascalvoc}, ours achieves state-of-the-art performance among models trained under the same level of supervision, and is competitive with those relying on stronger supervision or external data.
Surprisingly, it even outperforms FCN~\cite{Fcn}, the well-known fully supervised model in the early days.
\end{itemize}

% ======================================================================
% \recap{paper organization}
% ======================================================================
The rest of this paper is organized as follows. 
Section~\ref{sec:relatedwork} reviews previous approaches closely related to ours, and
Section~\ref{sec:method} describes each step of our framework in details.
Then we empirically evaluate the proposed framework on the public benchmark in Section~\ref{sec:experiments}, and conclude in Section~\ref{sec:conclusion} with brief remarks.

%%% RELATED WORK %%%%%%%%%%%%%%%%%%%%%%%%%%%%%%%%%%%%%%%%
% !TEX root = weaksup_jiwoon.tex

\section{Related Work}
\label{sec:relatedwork}

\noindent \textbf{Various types of weak supervision:}
Weakly supervised approaches have been extensively studied for semantic segmentation to address the data deficiency problem.
Successful examples of weak supervision for semantic segmentation include bounding box~\cite{Boxsup,wssl,KhorevaCVPR17}, scribble~\cite{scribblesup,Vernaza_2017_CVPR}, point~\cite{Bearman16}, and so on.
% , and natural language description~\cite{LinCVPR16}.
% They utilize labels that are weaker than the pixel-wise masks, but readily available in existing datasets or easily obtainable thanks to low annotation costs.
However, these types of weak supervision still require a certain amount of human intervention during annotation procedure, so it is costly to annotate these weak labels for a large number of visual data.

\noindent \textbf{Image-level labels as weak supervision:}
Image-level class labels have been widely used as weak supervision for semantic segmentation since they demand minimum or no human intervention to be annotated.
Early approaches have tried to train a segmentation model directly from image-level labels~\cite{wssl,Ccnn}, but their performance is not satisfactory since the labels are too coarse to teach segmentation. % by themselves
% Early approaches have tried to train a segmentation model directly from image-level labels by, for example, self-supervised learning~\cite{wssl} or employing a prior knowledge about object size as a constraint during training~\cite{Ccnn}.
% Unfortunately, performance of these approaches are not satisfactory since image-level labels are too coarse to teach segmentation by themselves.
To address this issue, some of previous arts incorporate segmentation seeds given by discriminative localization techniques~\cite{Cam,Oquab2014} with additional evidences like superpixels~\cite{KwakAAAI17,HongCVPR17,Wsl}, segmentation proposals~\cite{Wsl}, and motions in video~\cite{Tokmakov16,HongCVPR17}, which are useful to estimate object shapes and obtained by off-the-shelf unsupervised techniques.

Our framework based on AffinityNet has a clear advantage over the above approaches.
AffinityNet learns from data how to propagate local activations to entire object area, while the previous methods cannot take such an advantage.
% advantage of representation learning.
Like ours, a few methods improve segmentation quality without off-the-shelf preprocessing.
Wei~\etal~\cite{WeiCVPR17} propose to progressively expand segmentation results by searching for new and complementary object regions sequentially.
On the other hand, Kolesnikov and Lampert~\cite{sec} learn a segmentation model to approximate the output of dense Conditional Random Field (dCRF)~\cite{denseCRF} applied to the segmentation seeds given by CAMs.

\noindent \textbf{Learning pixel-level affinities:}
Our work is also closely related to the approaches that learn to predict affinity matrices in pixel-level~\cite{Bertasius_2017_CVPR,Vernaza_2017_CVPR,ChengCVPR2017}.
Specifically, a pixel-centric affinity matrix of an image is estimated by a DNN trained with segmentation labels in~\cite{Bertasius_2017_CVPR,ChengCVPR2017}.
% \jw{Maire~\etal~\cite{MaireCVPR2016} explored training a convolutional neural network to predict the affinity matrix for clustering pixels into regions with the purpose of image segmentation.}
Bertasius~\etal~\cite{Bertasius_2017_CVPR} incorporate the affinity matrix with random walk, whose role is to refine the output of a segmentation model like dCRF.
Cheng~\etal~\cite{ChengCVPR2017} design a deconvolution network, in which unpooling layers leverage the affinity matrix to recover sharp boundaries during upsampling.
Both of the above methods aim to refine outputs of fully supervised segmentation models in a pixel-level.
On the contrary, our goal is to recover object shape from coarse and noisy responses of object parts with a high-level semantic affinity matrix, and AffinityNet has a totally different architecture accordingly.
% On the contrary, the goal of our AffinityNet is to predict semantic affinity for revising rather incomplete initial segmentation in a weakly supervised setting, thus it has a totally different architecture accordingly.
Vernaza and Chandraker~\cite{Vernaza_2017_CVPR} employ scribbles as weak supervision, and propose to learn a segmentation network and the random walk affinity matrix simultaneously so that the output of the network and that of random walk propagation of the scribbles become identical.
Our approach is different from this work in the following three aspects.
First, our framework is trained with image-level labels, which are significantly weaker than the scribbles employed in~\cite{Vernaza_2017_CVPR}.
Second, in our approach random walk can jump to any other positions within a certain radius, but in~\cite{Vernaza_2017_CVPR} it is allowed to move only to four nearest neighbors.
Third, AffinityNet learns pairwise semantic affinity explicitly, but the model in~\cite{Vernaza_2017_CVPR} learns it implicitly.
% as the loss function is not directly related to the affinity.

\noindent \textbf{Learning with synthetic labels:}
We adopt the disjoint pipeline that first generates synthetic labels and train a segmentation model with the labels in a fully supervised manner. 
Such a pipeline has been studied for object detection~\cite{OICR} as well as semantic segmentation~\cite{Boxsup,KhorevaCVPR17,OhCVPR17,Tokmakov16,KwakAAAI17,HongCVPR17} in weakly supervised settings.
% It has been shown empirically in~\cite{KwakAAAI17,OhCVPR17} that the final models trained with synthetic labels outperforms those generating the labels.
A unique feature of our approach is the AffinityNet, an end-to-end trainable DNN improving the quality of synthetic labels significantly, when compared to previous methods that adopt existing optimization techniques (\eg, GraphCut, GrabCut, and dCRF) and/or off-the-shelf preprocessing steps aforementioned.

\iffalse
\noindent \textbf{Synthesizing segmentation labels:}
Although these methods can directly predict the segmentation masks from such weak labels, most approaches also have a separate segmentation network that is trained with the intermediate results as labels.
This learning scheme not only eliminates the need for labels except during training, but also improves final segmentation results further, as reported in~\cite{Boxsup, wssl, KhorevaCVPR17, OhCVPR17, Tokmakov16}.
In particular,~\cite{KhorevaCVPR17, OhCVPR17} employ the concept of ignore region, which helps training by not utilizing the uncertain areas of label maps.
\fi

%%% METHOD %%%%%%%%%%%%%%%%%%%%%%%%%%%%%%%%%%%%%%%%%%%%%%
% !TEX root = weaksup_jiwoon.tex

\section{Our Framework}
\label{sec:method}

% ======================================================================

% The overall pipeline of the proposed framework is illustrated in Figure~\ref{fig:outline}.
Our approach to weakly supervised semantic segmentation is roughly divided into two parts:
(1) Synthesizing pixel-level segmentation labels of training images given their image-level class labels, and
(2) Learning a DNN for semantic segmentation with the generated segmentation labels.
The entire framework is based on three DNNs:
A network computing CAMs, AffinityNet, and a segmentation model.
The first two are used to generate segmentation labels of training images, and the last one is the DNN that performs actual semantic segmentation and is trained with the synthesized segmentation annotations.
The remainder of this section describes characteristics of the three networks and training schemes for them in details.
% Specifications of the three networks and training schemes for them are described in the following sections.
% \suha{may need a bit more details about the entire framework andthe roles of the three DNNs?}

\subsection{Computing CAMs}
\label{sec:cam}
CAMs play an important role in our framework. 
As in many other weakly supervised approaches, they are considered as segmentation seeds, which typically highlight local salient parts of object and later propagate to cover the entire object area.
Furthermore, in our framework they are employed as sources of supervision for training AffinityNet.

\begin{figure} [!t]
\centering
\includegraphics[width = 0.95 \linewidth]{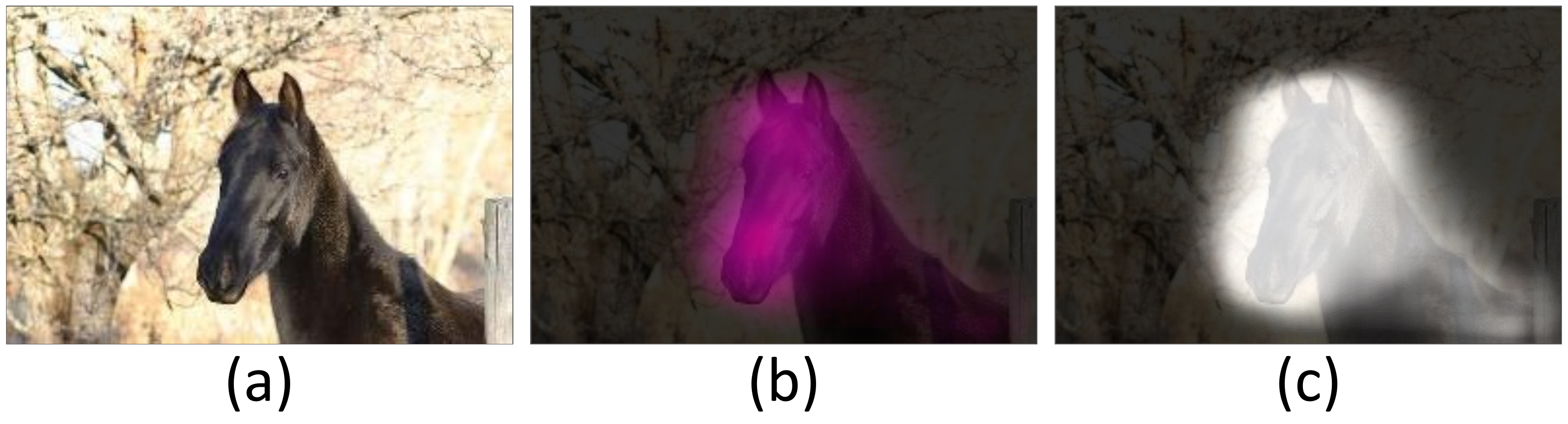}
\caption{Visualization of CAMs obtained by our approach.
(a) Input image. (b) CAMs of object classes: Brighter means more confident object region. (c) CAMs of background: Darker means more confident background region.} 
\label{fig:cams}
\end{figure}

We follow the approach of~\cite{Cam} to compute CAMs of training images.
The architecture is a typical classification network with global average pooling (GAP) followed by a fully connected layer, and is trained by a classification criteria with image-level labels.
Given the trained network, the CAM of a groundtruth class $c$, which is denoted by $M_c$, is computed by
\begin{equation}
    M_c (x, y) = \mathbf{w}_c^\top  f^\text{cam} (x, y),
\end{equation}
where $\mathbf{w}_c$ is the classification weights associated to the class $c$ and $f^\text{cam}(x,y)$ indicates the feature vector located at $(x,y)$ on the feature map before the GAP. % of the network.
$M_c$ is further normalized so that the maximum activation equals 1: $M_c (x, y) \rightarrow M_c (x, y) / \max_{x, y} {M_c (x, y)}$.
For any class $c'$ irrelevant to the groundtruths, we disregard $M_{c'}$ by making its activation scores zero.
We also estimate a background activation map, which is given by
\begin{equation}
    M_\text{bg} (x, y) = \Big\{ 1 - \max_{c \in C} {M_c (x, y)} \Big\}^\alpha \label{eq:cam_bg}
\end{equation}
where $C$ is the set of object classes and $\alpha \geq 1$ denotes a hyper-parameter that adjusts the background confidence scores.
Qualitative examples of CAMs obtained by our approach are visualized in Figure~\ref{fig:cams}.

%% ======================================================================
%% FIGURE START

\subsection{Learning AffinityNet}
\label{sec:learn_affinitynet}
%% ======================================================================
% \recap{what is AffinityNet}
% - goal
% - input and output
% - architecture: 
%  -- learning feature map, whose feature vectors meet the semantic distance relationship?
%% ======================================================================
AffinityNet aims to predict class-agnostic semantic affinity between a pair of adjacent coordinates on a training image.
The predicted affinities are used in random walk as transition probabilities so that random walk propagates activation scores of CAMs to nearby areas of the same semantic entity, which improves the quality of CAMs significantly.

\begin{figure} [!t]
\centering
\includegraphics[width = 0.95 \linewidth]{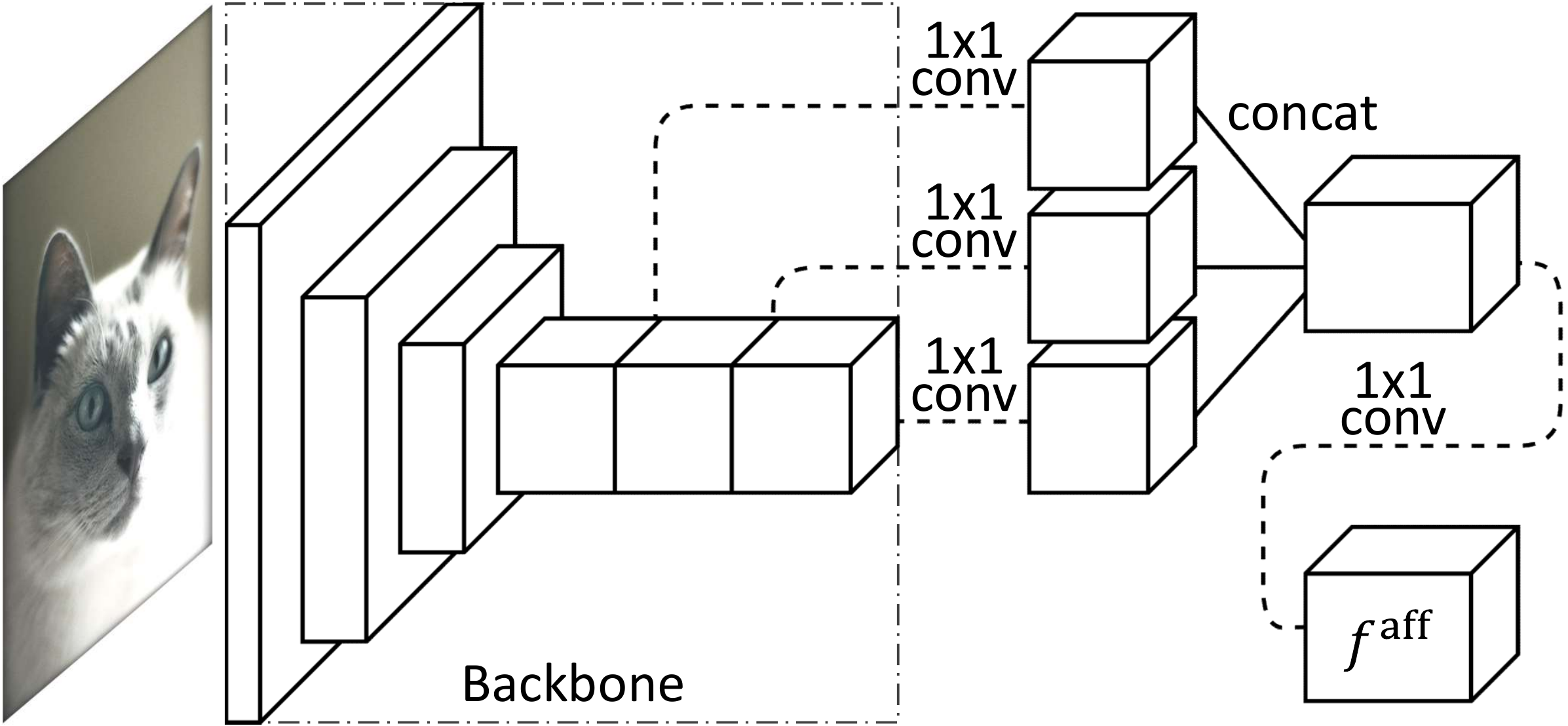}
\caption{
% \jw{Overall architecture of AffinityNet. 
% AffinityNet aims to process an image as convolutional features $f^\text{aff}$ that are used to measure affinity between coordinates in a pair.
% To enlarge the resolution of the feature maps, atrous convolution is adopted in the backbone network to increase receptive field size without downsampling. By skip-connections, the network aggregates various semantic contexts at multi-levels. The aggregated features are adapted to compute semantic affinities by 1$\times$1 convoltuions denoted by dashed lines. The details of backbone network and AffinityNet will be covered in section 4.1 and 4.2.}
Overall architecture of AffinityNet.
The output feature map $f^\text{aff}$ is obtained by aggregating feature maps from multiple levels of a backbone network so that $f^\text{aff}$ can take semantic information at various field-of-views.
Specifically, we first apply 1$\times$1 convolutions to the multi-level feature maps for dimensionality reduction, concatenate the results as a single feature map, and employ one more 1$\times$1 convolution for adaptation to the target task.
More details of the architecture is described in Section~\ref{sec:architecture}.
} 
\label{fig:affinitynet}
\end{figure}

For computational efficiency, AffinityNet is designed to predict a convolutional feature map $f^\text{aff}$ where the semantic affinity between a pair of feature vectors is defined in terms of their $L_1$ distance. 
Specifically, the semantic affinity between features $i$ and $j$ is denoted by $W_{i j}$ and defined as
\begin{equation}
    W_{i j} = \exp \Big\{ - \big\lVert f^\text{aff} (x_i, y_i) - f^\text{aff} (x_j, y_j) \big\rVert_1 \Big\}, \label{eq:def_affinity}
\end{equation}
where $(x_i,y_i)$ indicates the coordinate of the $i^\text{th}$ feature on the feature map $f^\text{aff}$. 
In this way, a large number of semantic affinities in a given image can be computed efficiently by a single forward pass of the network.
Figure~\ref{fig:affinitynet} illustrates the AffinityNet architecture and the way it computes $f^\text{aff}$.
Training this architecture requires semantic affinity labels for pairs of feature map coordinates, \ie, labels for $W_{ij}$ in Eq.~\eqref{eq:def_affinity}.
However, such labels are not directly available in our setting where only image-level labels are given.
%However, such labels are unavailable in our setting where only image-level labels are given.
% To train this architecture, one needs affinity labels for pairs of adjacent image coordinates, which is a strong supervision unavailable in our weakly supervised setting.
In the remaining part of this section, we present how to generate the affinity labels and train AffinityNet with them.

\iffalse
The architecture of AffinityNet is illustrated in Figure~\ref{fig:affinitynet}, and details of the network will be discussed in Section~\ref{sec:architecture}.
For computational efficiency, AffinityNet is designed to predict a convolutional feature map (\ie, $f^\text{aff}$ in Figure~\ref{fig:affinitynet}) where the semantic affinity between a pair of feature vectors is defined in terms of their $L_1$ distance. 
Specifically, the semantic affinity between features $i$ and $j$ is denoted by $W_{i j}$ and defined as
\begin{equation}
    W_{i j} = \exp \Big\{ - \big\lVert f^\text{aff} (x_i, y_i) - f^\text{aff} (x_j, y_j) \big\rVert_1 \Big\}, \label{eq:def_affinity}
\end{equation}
where $(x_i,y_i)$ indicates the coordinate of the $i^\text{th}$ feature on the feature map. 
In this way, a large number of semantic affinities in a given image can be computed efficiently by a single forward pass of the network.
\fi

%% ======================================================================
%% FIGURE START
%% FIGURE END
%% ======================================================================

\subsubsection{Generating Semantic Affinity Labels}
%% ======================================================================
% \recap{generating semantic affinity labels}
% - what do we need to train AffinityNet
%  -- a pair of adjacent image coordinates and a binary label for the pair
% - We employ CAMs as a source of supervision.
% - CAMs are not sufficiently accurate
% - but can obtain reliable supervision by
%  0) using CAMs of relevant classes only
%  1) varying the value of alpha
%  2) applying dCRF
% - see the figure, confident object/background regions + don'care part
% - we then sample pairs within a certain radius
% - we restrict the radius ... (from section 3.3)
%% ======================================================================
To train AffinityNet with image-level labels, we exploit CAMs of training images as incomplete sources of supervision.
Although CAMs are often inaccurate as shown in Figure~\ref{fig:cams}, we found that by carefully manipulating them, reliable supervision for semantic affinity can be obtained.
%Specifically, three cleaning steps are applied to CAMs of each training image before generating affinity labels. 
%First, the activation scores of classes irrelevant to the groundtruths are disregarded by setting their activation scores zero.
% Although CAMs are often inaccurate as shown in Figure~\ref{fig:cams}, we found that \jw{by selecting the confident parts of CAMs and then finding the semantic pair affinity among them, solid} supervision for learning semantic affinity can be obtained.

\begin{figure} [!t]
\centering
\includegraphics[width = 0.95 \linewidth]{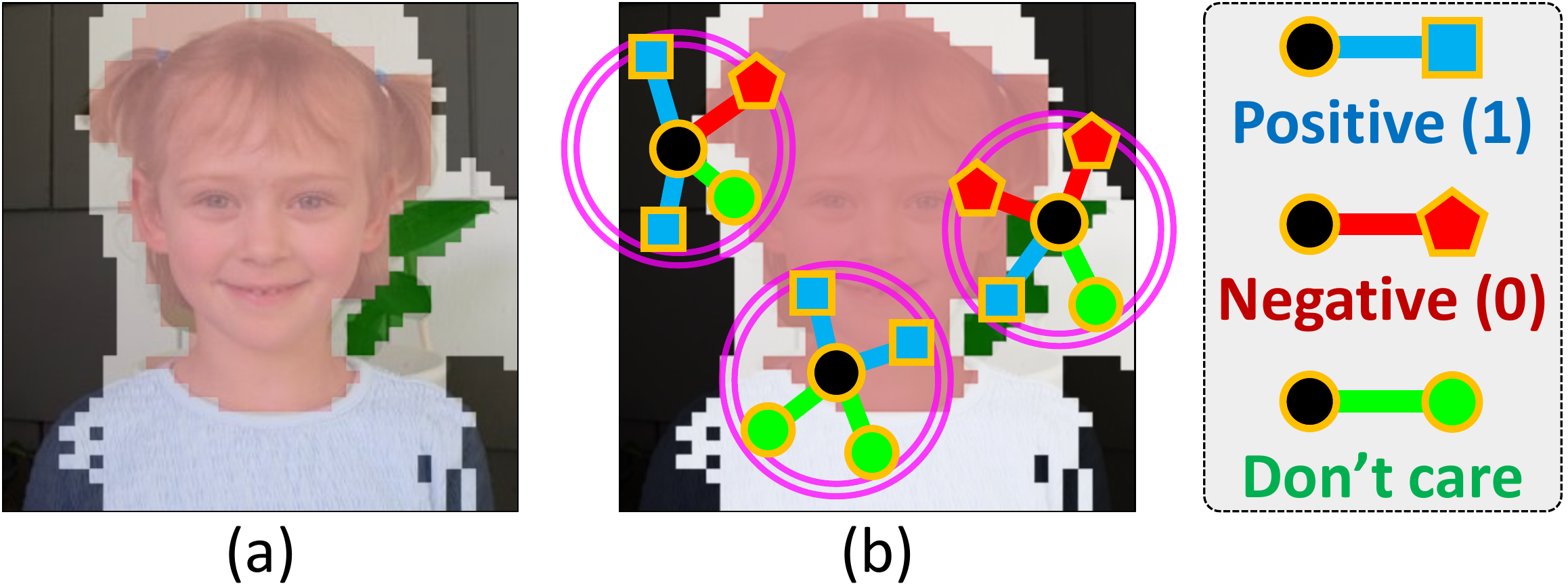}
\caption{
Conceptual illustration of generating semantic affinity labels.
(a) Confident areas of object classes and background: peach for \emph{person}, green for \emph{plant}, and black for \emph{background}. The \emph{neutral} area is color-coded in white. 
(b) Coordinate pairs sampled within a small radius for training AffinityNet. Each pair is assigned label 1 if its two coordinates come from the same class, and label 0 otherwise. When at least one of the two coordinates belongs to the \emph{neutral} area, the pair is ignored during training.
% (Best viewed in color) 
} 
\label{fig:pair}
\end{figure}

Our basic idea is to identify confident areas of objects and background from the CAMs, and sample training examples only from these areas.
By doing so, the semantic equivalence between a pair of sampled coordinates can be determined reliably.
To estimate confident areas of objects, we first amplify $M_\text{bg}$ by decreasing $\alpha$ in Eq.~\eqref{eq:cam_bg} so that background scores dominate insignificant activation scores of objects in the CAMs.
% After applying dCRF to the CAMs for their refinement, we identify confident object areas whose class activation scores are still greater than the amplified background scores.
% Also, by doing the opposite (\ie, increasing $\alpha$ to weaken $M_\text{bg}$), confident background areas are identified.
% Remaining areas in the image are then considered as \emph{neutral}.
% A result of this procedure is visualized in Figure~\ref{fig:pair}(a).
After applying dCRF to the CAMs for their refinement, we identify confident areas for each object class by collecting coordinates whose scores for the target class are greater than those of any other classes including the amplified background.
Also, in the opposite setting (\ie, increasing $\alpha$ to weaken $M_\text{bg}$), confident background areas can be identified in the same manner.
Remaining areas in the image are then considered as \emph{neutral}.
A result of this procedure is visualized in Figure~\ref{fig:pair}(a).

% Now a binary affinity label can be assigned to every pair of coordinates according to their class labels.
Now a binary affinity label can be assigned to every pair of coordinates according to their class labels determined by the confident areas.
For two coordinates $(x_i,y_i)$ and $(x_j,y_j)$ that are not \emph{neutral}, their affinity label $W^*_{ij}$ is 1 if their classes are the same, and 0 otherwise.
Also, if at least one of the coordinates is \emph{neutral}, we simply ignore the pair during training.
This scheme, which is illustrated in Figure~\ref{fig:pair}(b),  enables us to collect a fairly large number of pairwise affinity labels which are also reliable enough.

%% FIGURE END
%% ======================================================================

\subsubsection{AffinityNet Training}
%% ======================================================================
% \recap{AffinityNet training}
% - learning with affinity labels of local, spatially adjacent pairs
%   -- intuitively, if the distance between two are very far, semantic 
%      semantic affinity does not work: why? because context becomes weak. (lose context)
% - learning to make W_ij follows the groundtruth affinity label
%% ======================================================================

AffinityNet is trained by approximating the binary affinity labels $W^*_{ij}$ with the predicted semantic affinities $W_{ij}$ of Eq.~\eqref{eq:def_affinity} in a gradient descent fashion.
Especially, affinities of only sufficiently adjacent coordinates are considered during training due to the following two reasons.
First, it is difficult to predict semantic affinity between two coordinates too far from each other due to the lack of context.
Second, by addressing pairs of adjacent coordinates only, we can reduce computational cost significantly.
Thus the set of coordinate pairs used in training, denoted by $\mathcal{P}$, is given by
\begin{equation}
% \mathcal{P} = \Big\{(i, j) \Big\mid \sqrt{(x_i-x_j)^2 + (y_i - y_j)^2} < \gamma, i\neq j \Big\},
    \mathcal{P} = \big\{(i, j) \mid \mathrm{d}\big( (x_i, y_i), (x_j, y_j) \big) < \gamma, \forall i\neq j \big\} \label{eq:pair}
\end{equation}
where $\mathrm{d}(\cdot,\cdot)$ is the Euclidean distance and $\gamma$ is a search radius that limits the distance between a selected pair.

However, learning AffinityNet directly from $\mathcal{P}$ is not desirable due to the class imbalance issue.
We observed that in $\mathcal{P}$ the class distribution is significantly biased to positive ones since negative pairs are sampled only around object boundaries. %while a large number of positive pairs can be obtained from object and background areas.
Also in the subset of positive pairs, the number of background pairs is notably larger than that of object pairs as background is larger than object areas in many photos.
To address this issue, we divide $\mathcal{P}$ into three subsets, and aggregate losses obtained from individual subsets. % to address the class imbalance issue in $\mathcal{P}$. 
Specifically, we first divide $\mathcal{P}$ into two subsets of positive and negative pairs:
\begin{eqnarray}
    \mathcal{P}^+ & = & \big\{(i, j) \mid (i, j)\in \mathcal{P}, W^*_{ij} = 1 \big\}, \\
    \mathcal{P}^- & = & \big\{(i, j) \mid (i, j)\in \mathcal{P}, W^*_{ij} = 0 \big\},
\end{eqnarray}
and further break $\mathcal{P}^+$ into $\mathcal{P}^+_\text{fg}$ and $\mathcal{P}^+_\text{bg}$ for objects and background, respectively. 
Then the cross-entropy loss is computed per subset as follows:
\begin{eqnarray}
    \mathcal{L}^{+}_\text{fg} & = & - \frac{1}{| \mathcal{P}_\text{fg}^+ |} \sum_{(i, j)\in \mathcal{P}^+_\text{fg}} \log  W_{i j},  \\
    % encourages seed at the background to propagate
    \mathcal{L}^{+}_\text{bg} & = & - \frac{1}{| \mathcal{P}_\text{bg}^+ |} \sum_{(i, j)\in \mathcal{P}^+_\text{bg}} \log  W_{i j},  \\
    % encourages seed at the foreground to propagate
    \mathcal{L}^{-} & = & - \frac{1}{| \mathcal{P}^- |} \sum_{(i, j)\in \mathcal{P}^-} \log  ( 1 - W_{i j} ).
\end{eqnarray}
Finally, the loss for training the AffinityNet is defined as
\begin{equation}
    \mathcal{L} = \mathcal{L}_\text{fg}^{+} +  \mathcal{L}_\text{bg}^{+} + 2\mathcal{L}^{-}. \label{eqn:aff_loss}
\end{equation}
Note that the loss in Eq.~\eqref{eqn:aff_loss} is class-agnostic.
Thus the trained AffinityNet decides the class consistency between two adjacent coordinates while not aware of their classes explicitly.
This class-agnostic scheme allows AffinityNet to learn a more general representation that can be shared among multiple object classes and background, and enlarges the set of training samples per class significantly.
% For example, body parts of \emph{dog} and \emph{cat} are sometimes similar, and learning representation from one to apply another is a natural way to strengthen the representation power.
% Also, the missing class information is provided by CAMs.

\subsection{Revising CAMs Using AffinityNet}
\label{sec:revising_cams}

The trained AffinityNet is used to revise CAMs of training images.
Local semantic affinities predicted by AffinityNet are converted to a transition probability matrix, which enables random walk to be aware of semantic boundaries in image, 
and encourages it to diffuse activation scores within those boundaries.
We empirically found that random walk with the semantic transition matrix significantly improves the quality of CAMs and allows us to generate accurate segmentation labels consequently.

For an input image, AffinityNet generates a convolutional feature map, and semantic affinities between features on the map are computed according to Eq.~\eqref{eq:def_affinity}.
Note that, as in training of AffinityNet, the affinities are computed between features within local circles of radius $\gamma$.
The computed affinities form an affinity matrix $W$, whose diagonal elements are 1.
The transition probability matrix $T$ of random walk is derived from the affinity matrix as follows:
\begin{eqnarray}
    T = D^{-1} W^{\circ \beta}, \ \ \textrm{where} \ \ D_{i i} = \sum_j W_{i j}^{\beta}. \label{eq:trans_mat}
\end{eqnarray}
In the above equation, the hyper-parameter $\beta$ has a value greater than 1 so that $W^{\circ \beta}$, the Hadamard power of the original affinity matrix, ignores immaterial affinities in $W$. % and keeps those strong enough to survive after the Hadamard power.
Thus using $W^{\circ \beta}$ instead of $W$ makes our random walk propagation more conservative.
The diagonal matrix $D$ is computed for row-wise normalization of $W^{\circ \beta}$.

Through random walk with $T$, a single operation of the semantic propagation is implemented by multiplying $T$ to the CAMs.
We perform this propagation iteratively until the predefined number of iterations is reached.
Then $M^*_c$, the revised CAM of class $c$ is given by
\begin{equation}
   \text{vec}(M^*_c) = T^t \cdot \text{vec}(M_c) \quad \forall c\in C \cup \{ \text{bg} \}, \label{eq:rw_prop}
\end{equation}
where $\text{vec}(\cdot)$ means vectorization of a matrix, and $t$ is the number of iterations.
Note that the value of $t$ is set to a power of 2 so that Eq.~\eqref{eq:rw_prop} performs matrix multiplication only $\log_2 t + 1$ times.

\subsection{Learning a Semantic Segmentation Network}
\label{sec:learn_segnet}
% ======================================================================
% - how to generate segmentation labels from $M^*_c$
%   -- argmax per location
%   -- dCRF
% - segmentation network architecture: any type of segmentation network can be used.
% - difference between other approaches learning segmentation generator and segmentation network at the same time.
% ======================================================================

The revised CAMs of training images are then used to generate segmentation labels of the images.
Since CAMs are smaller than their input image in size, we upsample them to the resolution of the image by bilinear interpolation, and refine them with dCRF.
% A segmentation label of a training image is then obtained simply by selecting the label with maximum activation score in the revised CAMs at each location.
A segmentation label of a training image is then obtained simply by selecting the class label associated with the maximum activation score at every pixel in the revised and upsampled CAMs.
Note that the \emph{background} class can be also selected since we compute CAMs for background as well as object classes.
% Note that the selected labels can be \emph{background} label since we compute CAMs for background as well as object classes.

The segmentation labels obtained by the above procedure are used as supervision to train a segmentation network.
Any fully supervised semantic segmentation model can be employed in our approach as we provide segmentation labels of training images. 

% There are three reasons for learning an actual segmentation network, although CAMs and AffinityNet can be applied also to test images to obtain segmentation results.
% First, we can save computation time significantly by using a single network instead of the two networks (CAM and AffinityNet) and dCRF post-processing.
% Second, some techniques used to clean CAMs cannot be used for testing images whose groundtruth class labels are not available.
% Finally, by learning a segmentation network with synthesized segmentation labels, the segmentation performance increases as also reported in~\cite{KwakAAAI17,OhCVPR17}.

% ======================================================================
\section{Network Architectures}
\label{sec:architecture}

In this section, we present details of DNN architectures adopted in our framework.
Note that our approach can be implemented with any existing DNNs serving the same purposes, 
although we carefully design the following models to enhance segmentation performance.

\subsection{Backbone Network}
The three DNNs in our framework are all built upon the same backbone network.
The backbone is a modified version of Model A1~\cite{ZifengArxiv16}, which is also known as ResNet38 and has 38 convolution layers with wide channels.
To obtain the backbone network, the ultimate GAP and fully connected layer of the original model are first removed.
Then the convolution layers of the last three levels\footnote{A level is a group of residual units that share the same output stride.} are replaced by atrous convolutions with a common input stride of 1, and their dilation rates are adjusted so that the backbone network will return a feature map of stride 8.
The atrous convolution has been known to enhance segmentation quality by enlarging receptive field without sacrificing feature map resolution~\cite{ChenTPAMI17}.
We empirically observed that it works also in our weakly supervised models, CAM and AffinityNet, as it enables the models to recover fine shapes of objects.

\subsection{Details of DNNs in Our Framework}

% \paragraph{The network computing CAMs.}
\noindent \textbf{Network computing CAMs:}
We obtain this model by adding the following three layers on the top of the backbone network in the order:
a 3$\times$3 convolution layer with 512 channels for a better adaptation to the target task, a global average pooling layer for feature map aggregation, and a fully connected layer for classification.
% On the top of the backbone network, a $3\times3$ convolution layer is added for a better adaptation to the target task. Then we put a global average pooling layer for feature map aggregation, which is followed by a fully connected layer for classification.

% \paragraph{AffinityNet.}
\noindent \textbf{AffinityNet:}
%This network is basically designed to aggregate feature maps output by the last three atrous convolution layers of the backbone in order to take multi-level semantic information into account when computing affinities.
% \jw{This network is designed to aggregate feature maps with various field-of-views of the backbone network in order to take multi-level semantic information into account when computing affinities.}
This network is designed to aggregate multi-level feature maps of the backbone network in order to leverage semantic information acquired at various field-of-views when computing affinities.
%Before the aggregation, we reduce the dimensionalities of the three feature maps by passing them through individual $1\times1$ convolution layers. The reduced dimensions are 128, 256, and 512, for the first, second, and third feature maps, respectively.
% \jw{The backbone can be divided into 6 stages by the feature map size and the dilation rate. Each last convolution layer of the last three stages is taken through skip-connection, and reduce to 128, 256, and 512 channels respectively by passing them through individual $1\times1$ convolution layers.}
For the purpose, the feature maps output from the last three levels of the backbone network are selected. 
Before aggregation, their channel dimensionalities are reduced to 128, 256, and 512, for the first, second, and third feature maps, respectively, by individual 1$\times$1 convolution layers.
Then the feature maps are concatenated to be a single feature map with 896 channels. 
We finally add one more 1$\times$1 convolution layer with 896 channels on the top for adaptation.

% \paragraph{The segmentation model.}
\noindent \textbf{Segmentation model:}
We strictly follow~\cite{ZifengArxiv16} to build our segmentation network.
Specifically, we put two more atrous convolution layers on the top of the backbone.
%They have the same dilation rate of 12, but their channel numbers are different: 512 for the first one, and 21 for the second.
They have the same dilation rate of 12, while the numbers of channels are 512 for the first one and 21 for the second.
% The resulting network can be considered as a variant of DeepLab-LargeFOV~\cite{Deeplabcrf}.
The resulting network is called ``Ours-ResNet38'' in the next section.

% \jw{Like most competitors, we also test our segmentation model with VGG16 based DeepLab-CRF-LargeFOV~\cite{Deeplabcrf} on publicly available caffe framework~\cite{caffe}. In the rest of the paper, we simply denote the results of this model as Ours-DeepLab.} 

%%% EXPERIMENTS %%%%%%%%%%%%%%%%%%%%%%%%%%%%%%%%%%%%%%%%%
% !TEX root = weaksup_jiwoon.tex

\section{Experiments}
\label{sec:experiments}

%% ======================================================================
%% TABLE START
\begin{table}[!t] \small
\centering
\begin{tabular}{c|c||c|c|c}
    \hline
    \multicolumn{2}{c||}{Kwak~\etal~\cite{KwakAAAI17}} & \multicolumn{3}{c}{Ours}\\
    \hline
    CAM & SPN & CAM & CAM+RW & CAM+RW+dCRF \\
    \hline
    30.5 & 43.8 & 48.0 & 58.1 & 59.7 \\
    \hline
\end{tabular}
\vspace{0.1cm}
\caption{Accuracy of synthesized segmentation labels in mIoU, evaluated on the PASCAL VOC 2012 \emph{training} set.
SPN: Superpixel Pooling Net of~\cite{KwakAAAI17}, RW: random walk with AffinityNet.}
\label{tab:synth_anno_acc}
\end{table}
%% TABLE END
%% ======================================================================

%% ======================================================================
%% FIGURE START
\begin{figure} [!t]
\centering
\includegraphics[width = \linewidth]{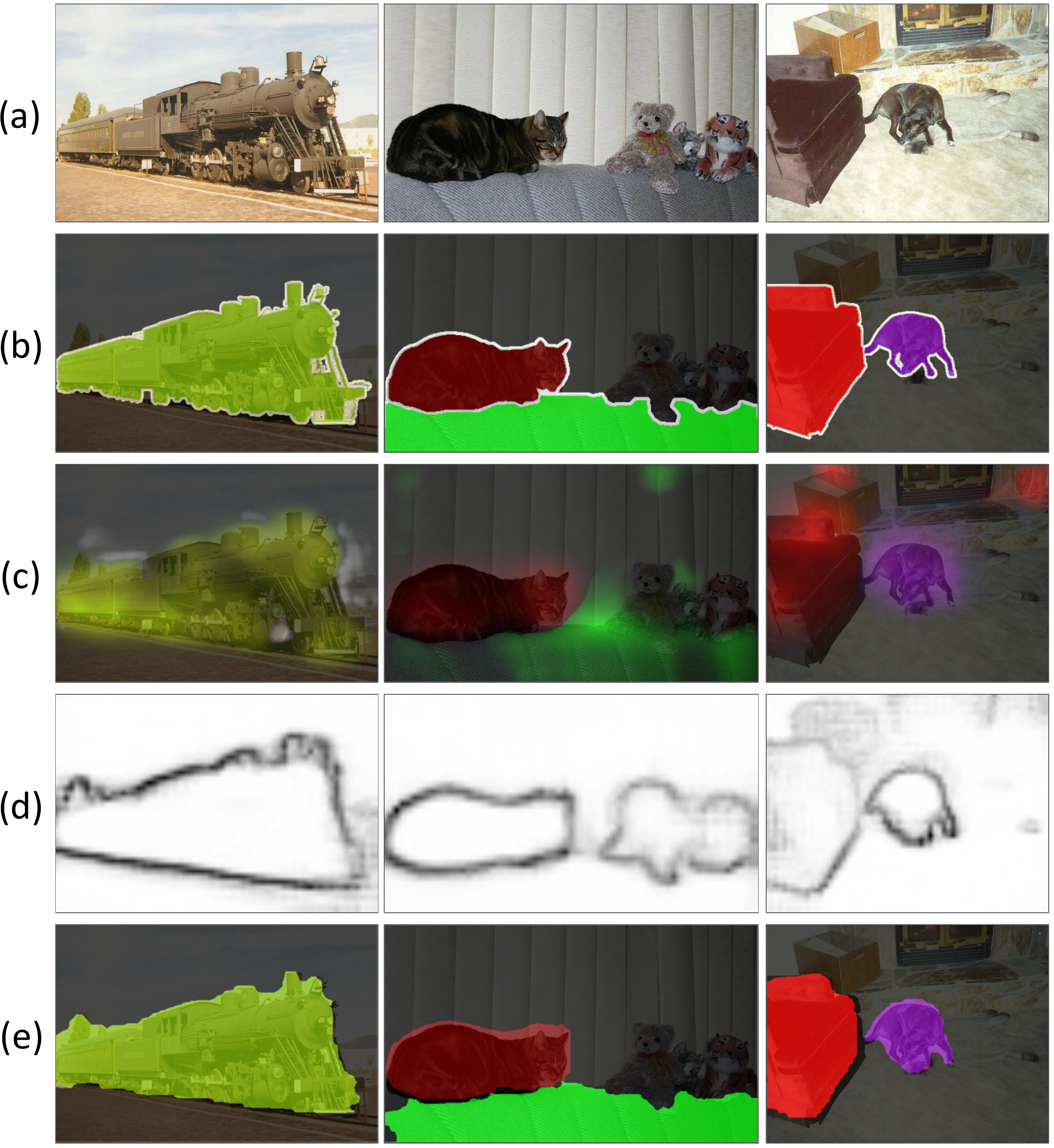}
\caption{Qualitative examples of synthesized segmentation labels of training images in the PASCAL VOC 2012 benchmark. (a) Input images. (b) Groundtruth segmentation labels. (c) CAMs of object classes. (d) Visualization of the predicted semantic affinities. (e) Synthesized segmentation annotations.
% without post-processing by dense CRF. 
% Semantic affinities predicted by our AffinityNet allow a simple random walk algorithm to recover object areas missing in the seeds and suppress false alarms effectively.
}
\label{fig:synth_anno}
\end{figure}
%% FIGURE END
%% ======================================================================

%% ======================================================================
%% TABLE START

\begin{table*}[!t] \footnotesize
\centering
\begin{tabular}
{
@{}C{2.0cm}@{}|@{}C{0.69cm}@{}C{0.67cm}@{}C{0.67cm}@{}C{0.67cm}@{}C{0.67cm}@{}C{0.67cm}@{}C{0.67cm}@{}C{0.67cm}@{}C{0.67cm}@{}C{0.67cm}@{}C{0.67cm}@{}C{0.67cm}@{}C{0.67cm}@{}C{0.67cm}@{}C{0.76cm}@{}C{0.76cm}@{}C{0.67cm}@{}C{0.67cm}@{}C{0.67cm}@{}C{0.67cm}@{}C{0.67cm}@{}|@{}C{0.76cm}@{}
}
\hline
Method\raggedright&bkg&aero&bike&bird&boat&bottle&bus&car&cat&chair&cow&table&dog&horse&mbk&person&plant&sheep&sofa&train&tv&mean\\
\hline
EM-Adapt~\cite{wssl} \raggedright & 67.2 & 29.2 & 17.6 & 28.6 & 22.2 & 29.6 & 47.0 & 44.0 & 44.2 & 14.6 & 35.1 & 24.9 & 41.0 & 34.8 & 41.6 & 32.1 & 24.8 & 37.4 & 24.0 & 38.1 & 31.6 & 33.8 \\ 
CCNN~\cite{Ccnn} \raggedright & 68.5 & 25.5 & 18.0 & 25.4 & 20.2 & 36.3 & 46.8 & 47.1 & 48.0 & 15.8 & 37.9 & 21.0 & 44.5 & 34.5 & 46.2 & 40.7 & 30.4 & 36.3 & 22.2 & 38.8 & 36.9 & 35.3 \\ 
MIL+seg~\cite{Wsl} \raggedright & 79.6 & 50.2 & 21.6 & 40.9 & 34.9 & 40.5 & 45.9 & 51.5 & 60.6 & 12.6 & 51.2 & 11.6 & 56.8 & 52.9 & 44.8 & 42.7 & 31.2 & 55.4 & 21.5 & 38.8 & 36.9 & 42.0 \\ 
SEC~\cite{sec} \raggedright & 82.4& 62.9& 26.4& 61.6& 27.6& 38.1& 66.6& 62.7& \bf{75.2}& 22.1& 53.5& 28.3& 65.8& 57.8& 62.3& 52.5& 32.5& 62.6& 32.1& 45.4& 45.3& 50.7 \\
AdvErasing~\cite{WeiCVPR17} \raggedright & 83.4& \bf{71.1}& 30.5& 72.9& 41.6& 55.9& 63.1& 60.2& 74.0& 18.0& \bf{66.5}& 32.4& 71.7& 56.3& 64.8& 52.4& 37.4& 69.1& 31.4& 58.9& 43.9 &55.0 \\
\hline
\bf{Ours-DeepLab} \raggedright & 87.2 & 57.4 & 25.6 & 69.8 & 45.7 & 53.3 & 76.6 & \bf{70.4} & 74.1 & 28.3 & 63.2 & \bf{44.8} & 75.6 & \bf{66.1} & 65.1 & 71.1 & 40.5 & 66.7 & 37.2 & 58.4 & 49.1 & 58.4 \\
\bf{Ours-ResNet38} \raggedright & \bf{88.2}& 68.2& \bf{30.6}& \bf{81.1}& \bf{49.6}& \bf{61.0}& \bf{77.8}& 66.1& 75.1& \bf{29.0}& 66.0& 40.2& \bf{80.4}& 62.0& \bf{70.4}& \bf{73.7}& \bf{42.5}& \bf{70.7}& \bf{42.6}& \bf{68.1}& \bf{51.6}& \bf{61.7}  \\
\hline
\end{tabular}
\vspace{0.1cm}
\caption{Performance on the PASCAL VOC 2012 \textit{val} set, compared to weakly supervised approaches based only on image-level labels.}
\label{tab:voc_result_val}
\vspace{-0.1cm}
\end{table*}

%% TABLE END
%% ======================================================================

%% ======================================================================
%% TABLE START

\begin{table*}[!t] \footnotesize
\centering
\begin{tabular}
{
@{}C{2.0cm}@{}|@{}C{0.69cm}@{}C{0.67cm}@{}C{0.67cm}@{}C{0.67cm}@{}C{0.67cm}@{}C{0.67cm}@{}C{0.67cm}@{}C{0.67cm}@{}C{0.67cm}@{}C{0.67cm}@{}C{0.67cm}@{}C{0.67cm}@{}C{0.67cm}@{}C{0.67cm}@{}C{0.76cm}@{}C{0.76cm}@{}C{0.67cm}@{}C{0.67cm}@{}C{0.67cm}@{}C{0.67cm}@{}C{0.67cm}@{}|@{}C{0.76cm}@{}
}
\hline
Method\raggedright&bkg&aero&bike&bird&boat&bottle&bus&car&cat&chair&cow&table&dog&horse&mbk&person&plant&sheep&sofa&train&tv&mean\\
\hline
EM-Adapt~\cite{wssl} \raggedright & 76.3 & 37.1 & 21.9 & 41.6 & 26.1 & 38.5 & 50.8 & 44.9 & 48.9 & 16.7 & 40.8 & 29.4 & 47.1 & 45.8 & 54.8 & 28.2 & 30.0 & 44.0 & 29.2 & 34.3 & 46.0 & 39.6 \\
CCNN~\cite{Ccnn} \raggedright & 70.1 & 24.2 & 19.9 & 26.3 & 18.6 & 38.1 & 51.7 & 42.9 & 48.2 & 15.6 & 37.2 & 18.3 & 43.0 & 38.2 & 52.2 & 40.0 & 33.8 & 36.0 & 21.6 & 33.4 & 38.3 & 35.6 \\
MIL+seg~\cite{Wsl} \raggedright & 78.7 & 48.0 & 21.2 & 31.1 & 28.4 & 35.1 & 51.4 & 55.5 & 52.8 & 7.8 & 56.2 & 19.9 & 53.8 & 50.3 & 40.0 & 38.6 & 27.8 & 51.8 & 24.7 & 33.3 & 46.3 & 40.6 \\
SEC~\cite{sec} \raggedright & 83.5& 56.4& 28.5& 64.1& 23.6& 46.5& 70.6& 58.5& 71.3& 23.2& 54.0& 28.0& 68.1& 62.1& 70.0& 55.0& 38.4& 58.0& 39.9& 38.4& 48.3& 51.7\\
AdvErasing~\cite{WeiCVPR17} \raggedright & -& -& -& -& -& -& -& -& -& -& -& -& -& -& -& -& -& -& -& -& -& 55.7 \\
\hline
\bf{Ours-DeepLab} \raggedright & 88.0& 61.1& 29.2& 73.0& 40.5& 54.1& 75.2& \bf{70.4}& \bf{75.1}& 27.8& 62.5& 51.4& 78.4& \bf{68.3}& 76.2& 71.8& 40.7& \bf{74.9}& \bf{49.2}& 55.0& 48.3& 60.5 \\  
\bf{Ours-ResNet38} \raggedright & \bf{89.1}& \bf{70.6}& \bf{31.6}& \bf{77.2}& \bf{42.2}& \bf{68.9}& \bf{79.1}& 66.5& 74.9& \bf{29.6}& \bf{68.7}& \bf{56.1}& \bf{82.1}& 64.8& \bf{78.6}& \bf{73.5}& \bf{50.8}& 70.7& 47.7& \bf{63.9}& \bf{51.1}& \bf{63.7} \\
\hline
\end{tabular}
\vspace{0.1cm}
\caption{Performance on the PASCAL VOC 2012 \textit{test} set, compared to weakly supervised approaches based only on image-level labels.}
\label{tab:voc_result_test}
\vspace{-0.1cm}
\end{table*}

%% TABLE END
%% ======================================================================

%% ======================================================================
%% TABLE START

\begin{table}[!t] \footnotesize
\centering
\begin{tabular}{lcc|cc}
\hline
Method & Sup. & Extra Data & \emph{val} & \emph{test} \\
\hline
TransferNet~\cite{transfernet} \raggedright  & $\mathcal{I}$ & MS-COCO~\cite{Mscoco}          & 52.1 & 51.2 \\
Saliency~\cite{OhCVPR17} \raggedright        & $\mathcal{I}$ & MSRA~\cite{LiuCVPR07}, BSDS~\cite{BSDS} & 55.7 & 56.7 \\
MCNN~\cite{Tokmakov16} \raggedright          & $\mathcal{I}$ & YouTube-Object~\cite{youtube-obj}  & 38.1 & 39.8 \\
CrawlSeg~\cite{HongCVPR17} \raggedright      & $\mathcal{I}$ & YouTube Videos     & 58.1 & 58.7 \\
\hline
What'sPoint~\cite{Bearman16} \raggedright    & $\mathcal{P}$ & - & 46.0 & 43.6 \\
RAWK~\cite{Vernaza_2017_CVPR} \raggedright   & $\mathcal{S}$ & - & 61.4 & -    \\
ScribbleSup~\cite{scribblesup} \raggedright  & $\mathcal{S}$ & - & 63.1 & -    \\
WSSL~\cite{wssl} \raggedright                & $\mathcal{B}$ & - & 60.6 & 62.2 \\
BoxSup~\cite{Boxsup} \raggedright            & $\mathcal{B}$ & - & 62.0 & 64.6 \\
SDI~\cite{KhorevaCVPR17} \raggedright        & $\mathcal{B}$ & BSDS~\cite{BSDS} & 65.7 & 67.5 \\
\hline
FCN~\cite{Fcn} \raggedright                  & $\mathcal{F}$ & - & -    & 62.2 \\
DeepLab~\cite{Deeplabcrf} \raggedright & $\mathcal{F}$ & - & 67.6 & 70.3 \\
ResNet38~\cite{ZifengArxiv16} \raggedright & $\mathcal{F}$ & - & 80.8 & 82.5 \\
\hline
\bf{Ours-DeepLab} \raggedright                       & $\mathcal{I}$ & - & 58.4 & 60.5 \\
\bf{Ours-ResNet38} \raggedright                       & $\mathcal{I}$ & - & 61.7 & 63.7 \\
\hline
\end{tabular}
\vspace{0.1cm}
\caption{Performance on the PASCAL VOC 2012 \emph{val} and \emph{test} sets. The supervision types (Sup.) indicate: $\mathcal{P}$--point, $\mathcal{S}$--scribble, $\mathcal{B}$--bounding box, $\mathcal{I}$--image-level label, and $\mathcal{F}$--segmentation label.}
\label{tab:comparison_suptype}
\vspace{-0.1cm}
\end{table}

%% TABLE END
%% ======================================================================

This section demonstrates the effectiveness of our approach with comparisons to current state of the art in weakly supervised semantic segmentation on the PASCAL VOC 2012 segmentation benchmark~\cite{Pascalvoc}. 
For a performance metric, we adopt Intersection-over-Union (IoU) between groundtruth and predicted segmentation.
% we adopt mean Intersection-over-Union (mIoU) between groundtruth labels and predicted segmentation masks.
% As an evaluation metric, we adopt Intersection-over-Union (IoU) between groundtruth and predicted segmentation.

\subsection{Implementation Details} \label{sec:exp_setup}
% \paragraph{Dataset.} 
\noindent \textbf{Dataset:}
All DNNs in our framework are trained and evaluated on the PASCAL VOC 2012 segmentation benchmark, for a fair comparison to previous approaches.
Following the common practice, we enlarge the set of training images by adopting segmentation annotations presented in~\cite{HariharanICCV2011}.
% augment segmentation annotations for training by adopting those presented in~\cite{HariharanICCV2011} to enlarge the set of training images.
Consequently 10,582 images in total are used as training examples and 1,449 images are kept for validation.

% \paragraph{Network parameter optimization.}
\noindent \textbf{Network parameter optimization:}
% Our approach is implemented in Tensorflow~\cite{tensorflow}. 
The backbone network of our DNNs is pretrained on the ImageNet~\cite{Imagenet}.
 % before finetuned on the PASCAL VOC 2012. 
The entire network parameters are then finetuned on the PASCAL VOC 2012 by Adam~\cite{Adamsolver}.
 % with the initial learning rate 0.001 and default parameters. 
%When learning the DNNs, a few data augmentation techniques are commonly used: horizontal flip, random cropping, random lightening~\cite{Alexnet}, and random shift saturation and contrast \suha{could you please check their names once again? also, say ``random'' once.}.
% When learning the DNNs, the following data augmentation techniques are commonly used: horizontal flip, random cropping, and color jittering~\cite{Alexnet}.
The following data augmentation techniques are commonly used when training all the three DNNs: horizontal flip, random cropping, and color jittering~\cite{Alexnet}.
%Particularly for the DNN computing CAMs, we randomly scale input images during training, which is useful to impose scale invariance on the network~\cite{HongCVPR17}.
% Particularly for the \jw{DNNs computing CAMs and semantic segmentation}, we randomly scale input images during training, which is useful to impose scale invariance on the network~\cite{HongCVPR17}.
Also, for the networks except AffinityNet, we randomly scale input images during training, which is useful to impose scale invariance on the networks.

% In the case of SeedNet, the data augmentation techniques enumerated above improves the quality of segmentation seeds significantly (7.2\% in mIoU). --> I removed it as measuring segmentation quality of CAMs is not straightforward...

% \suha{may we have to justify the advantages of the jittering techniques? probably not. do it only if we have enough tiem.}
% [We can train SeedNet without scaling augmentation and then measure mIoU score. But I think it is common technique in many classification network. But averaging the activation maps of multiscale gives huge gains. I've just measured mIoU score from our SeedNet without synthesizing multi-scale and flip images. And its mIoU is 40.725. (around 7.2 improvements when applying 20 images merging)]
% --> thanks!

% \paragraph{Hyper-parameter setting.} 
\noindent \textbf{Parameter setting:} 
$\alpha$ in Eq.~\eqref{eq:cam_bg} is 16 by default, and changed to 4 and 24 to amplify and weaken background activations, respectively.
We set $\gamma$ in Eq.~\eqref{eq:pair} to 5 and $\beta$ in Eq.~\eqref{eq:trans_mat} to 8.
Also, $t$ in Eq.~\eqref{eq:rw_prop} is fixed by 256.
For dCRF, we used the default parameters given in the original code.

\begin{figure*}[!ht]
\begin{center}
\includegraphics[width=\linewidth] {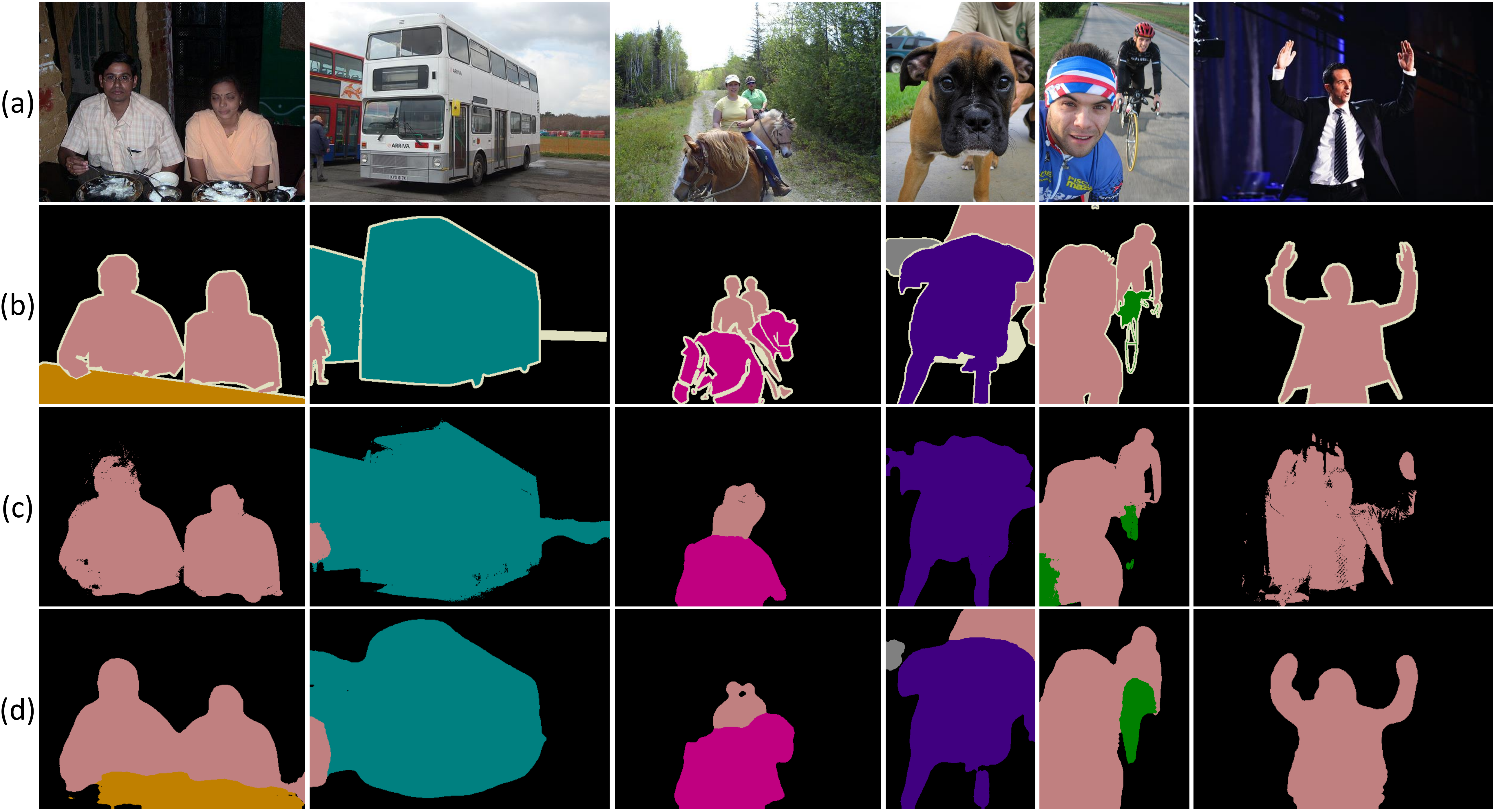}
\end{center}
\caption{
Qualitative results on the PASCAL VOC 2012 \emph{val} set. 
(a) Input images. (b) Groundtruth segmentation. (c) Results obtained by CrawlSeg~\cite{HongCVPR17}. (d) Results of Ours-ResNet38.
Compared to CrawlSeg, which is the current state-of-the-art model based on image-level label supervision, our method better captures larger object areas and less prone to miss objects. 
The object boundaries of our results are smoother than those of CrawlSeg as we do not apply dCRF to the final results.
More results can be found in the supplementary material.
} 
\label{fig:qualitative_voc}
\end{figure*}
%% FIGURE END
%% ======================================================================

\subsection{Analysis of Synthesized Segmentation Labels}

The performance of our label synthesis method is measured in mIoU between groundtruth and generated segmentation labels as summarized in Table~\ref{tab:synth_anno_acc}.
For ablation study, our method is divided into three parts: CAM, RW (random walk with AffinityNet), and dCRF.
To demonstrate the advantage of the proposed method, we also report the score of Superpixel Pooling Net (SPN)~\cite{KwakAAAI17} that incorporates CAM with superpixels as additional clues to generate segmentation labels with image-level label supervision.
% quality of segmentation seeds obtained by the original CAM~\cite{Cam}, and the score is adopted from~\cite{KwakAAAI17} in which the original network of~\cite{Cam} is directly adopted and finetuned on the PASCAL VOC 2012 with scale jittering.
% In the table, we also report the quality of segmentation seeds obtained by the original CAM~\cite{Cam} to demonstrate the advantage of our SeedNet.
% Specifically, for the original CAM, we directly adopt the network of~\cite{Cam} trained on the ImageNet and finetune its parameters on the PASCAL VOC 2012 with the data augmentation techniques used in training of our SeedNet.
As shown in Table~\ref{tab:synth_anno_acc}, even our CAM outperforms SPN in terms of the quality of generated segmentation labels without using off-the-shelf techniques like superpixel.
We believe this is because of the various data augmentation techniques and the more powerful backbone network with atrous convolution layers.
% ----------
% 1) Most importantly, our backbone network is pretty powerful. its performance measured in ImangeNet is even better than ResNeXT. 
% 2) Our backbone network get huge benefits from adopting atrous convolutions. As reported in Wider/Deeper paper, this network with atrous conv is more efficient than ResNet-101 with atrous conv.
% 3) strong data augmentations. color, flip, scale, cropping. 
% 4) better multiscale synthesizing practices: when merging the outputs average pooling is much better than max pooling as I showed you. And I've merged 20 images for one result which is pretty time comsuming. I've given more weights on higher resoltuion. (by multiplying scale) 
% 5) I lowered the threshold. Original cam suggests thresholding bottom 20\% score. The results of this may look nice. But its mIoU score is not good. Applying 0.125 thresholding (and this is almost same as applying alpha=16) gives huge mIoU gains but it has too much false positives and may not look nice to see. 
% -->> thanks for your comments. I guess #5 may not actually increases the quality of segmentation seeds (although it improves mIoU a lot, as it simply increases recall scores of 20 object classes while decreases that of background---please note that the number of pixels belonging to background dominates others, but mIoU simply average all the mIoU scores computed per class.,
% ----------
Moreover, through random walk with the learned semantic affinity, the quality of segmentation annotations is improved remarkably, which demonstrates the effectiveness of AffinityNet.
Finally, dCRF further improves the label quality slightly, and we employ this last version as the supervision for learning the segmentation network.

Examples of the synthesized segmentation labels are shown in Figure~\ref{fig:synth_anno}, where one can see that random walk with AffinityNet handles false positives and missing areas in CAMs effectively. 
% Figure~\ref{fig:synth_anno} shows examples of the synthesized segmentation labels.
% Random walk with AffinityNet handles false positives and missing areas in the segmentation seeds simply yet effectively. %, and recover accurate shapes of target objects consequently. 
To illustrate the role of AffinityNet in this process, we also visualized predicted semantic affinities of the images by detecting edges on the feature map $f^\text{aff}$, and observed that AffinityNet has capability to detect semantic boundaries although it is trained with image-level labels.
% , where one can see that the AffinityNet has a capability to recognize semantic boundaries in image although it is trained with image-level class labels only.
As such boundaries penalize random walk propagation between semantically different objects, the synthesized segmentation labels can recover accurate shapes of objects.

\subsection{Comparisons to Previous Work}

We first quantitatively compare our approach with previous methods based only on image-level class labels.
The results on the PASCAL VOC 2012 are summarized in Table~\ref{tab:voc_result_val} and~\ref{tab:voc_result_test}.
Note that we also evaluate DeepLab~\cite{ChenTPAMI17} trained with our synthetic labels, called ``Ours-DeepLab'', for fair comparisons to other models whose backbones are VGG16~\cite{Vgg16}.
% In the \emph{val} set of the benchmark, Ours-ResNet38 achieves the best for 18 categories among 21 including \emph{background}, and outperforms the current state of the art~\cite{WeiCVPR17} by a large margin in terms of mean accuracy.
Both of our models outperform the current state of the art~\cite{WeiCVPR17} by large margins in terms of mean accuracy on both \emph{val} and \emph{test} sets of the benchmark, while Ours-ResNet38 is slightly better than Ours-DeepLab thanks to the more powerful representation of ResNet38.
% Especially, noticeable improvements were made by our approach in \emph{bus}, \emph{sofa}, and \emph{train} classes which typically have repetitive patterns in appearance 
% chair person training images containing those classes 
% The same tendency was observed from the results on the PASCAL VOC 2012 \emph{test} set as summarized in Table~\ref{tab:voc_result_test}.
Our models are also compared to the approaches based on extra training data or stronger supervision in Table~\ref{tab:comparison_suptype}.
They substantially outperform the approaches based on the same level of supervision but with extra data and annotations like segmentation labels in MS-COCO~\cite{Mscoco}, class-agnostic bounding boxes in MSRA Saliency~\cite{LiuCVPR07}, and YouTube videos~\cite{youtube-obj}.
They are also competitive with previous arts relying on stronger supervision like scribble and bounding box. 
Surprisingly, Ours-ResNet38 outperforms even FCN~\cite{Fcn}, the well-known early work on fully supervised semantic segmentation.
These results show that segmentation labels generated by our method are sufficiently strong, and can substitute for extra data or stronger supervision.
We finally compare our models with their fully supervised versions, DeepLab~\cite{ChenTPAMI17} and ResNet38~\cite{ZifengArxiv16}, which are the upper bounds we can achieve.
Specifically, Ours-DeepLab recovers 86\% of its bound, and Ours-ResNet38 achieves 77\%.

Figure~\ref{fig:qualitative_voc} presents qualitative results of Ours-ResNet38 and compares them to those of CrawlSeg~\cite{HongCVPR17}, which is the current state of the art using image-level supervision.
Our method relying only on image-level label supervision tends to produce more accurate results even though CrawlSeg exploits extra video data to synthesize segmentation labels.

% our scores are impressive when regarding those of fully supervised techniques proposed in early days (\eg, hypercolumn and FCN); our approach outperforms hypercolumn and is competitive with FCN.

%%% CONCLUSION %%%%%%%%%%%%%%%%%%%%%%%%%%%%%%%%%%%%%%%%%%
% !TEX root = weaksup_jiwoon.tex

\section{Conclusion}
\label{sec:conclusion}

To alleviate the lack of annotated data issue in semantic segmentation,
we have proposed a novel framework based on AffinityNet to generate accurate segmentation labels of training images given their image-level class labels only.
% The key component of the approach is AffinityNet, which enables a simple random walk algorithm 
The effectiveness of our approach has been demonstrated on the PASCAL VOC 2012 benchmark, where DNNs trained with the labels generated by our method substantially outperform the previous state of the art relying on the same level of supervision, and is competitive with those demanding stronger supervision or extra data.

% \vspace{-0.4cm}
\paragraph{Acknowledgement:} This work was supported by Kakao, Korea Creative Content Agency (KOCCA), and Ministry of Culture, Sports, and Tourism (MCST) of Korea.

% \newpage
{\small
\bibliographystyle{ieee}

\begin{thebibliography}{10}\itemsep=-1pt

\bibitem{Bearman16}
A.~Bearman, O.~Russakovsky, V.~Ferrari, and L.~Fei-Fei.
\newblock {What's the Point: Semantic Segmentation with Point Supervision}.
\newblock In {\em Proceedings of the European Conference on Computer Vision
  (ECCV)}, pages 549--565, 2016.

\bibitem{Bertasius_2017_CVPR}
G.~Bertasius, L.~Torresani, S.~X. Yu, and J.~Shi.
\newblock Convolutional random walk networks for semantic image segmentation.
\newblock In {\em Proceedings of the IEEE Conference on Computer Vision and
  Pattern Recognition (CVPR)}, July 2017.

\bibitem{Deeplabcrf}
L.-C. Chen, G.~Papandreou, I.~Kokkinos, K.~Murphy, and A.~L. Yuille.
\newblock Semantic image segmentation with deep convolutional nets and fully
  connected {CRFs}.
\newblock In {\em Proceedings of the International Conference on Learning
  Representations (ICLR)}, 2015.

\bibitem{ChenTPAMI17}
L.~C. Chen, G.~Papandreou, I.~Kokkinos, K.~Murphy, and A.~L. Yuille.
\newblock Deeplab: Semantic image segmentation with deep convolutional nets,
  atrous convolution, and fully connected crfs.
\newblock {\em IEEE Transactions on Pattern Analysis and Machine Intelligence
  (TPAMI)}, PP(99):1--1, 2017.

\bibitem{ChengCVPR2017}
Y.~Cheng, R.~Cai, Z.~Li, X.~Zhao, and K.~Huang.
\newblock Locality-sensitive deconvolution networks with gated fusion for rgb-d
  indoor semantic segmentation.
\newblock In {\em Proceedings of the IEEE Conference on Computer Vision and
  Pattern Recognition (CVPR)}, July 2017.

\bibitem{Boxsup}
J.~Dai, K.~He, and J.~Sun.
\newblock {BoxSup}: Exploiting bounding boxes to supervise convolutional
  networks for semantic segmentation.
\newblock In {\em Proceedings of the IEEE International Conference on Computer
  Vision (ICCV)}, pages 1635--1643, 2015.

\bibitem{Imagenet}
J.~Deng, W.~Dong, R.~Socher, L.-J. Li, K.~Li, and L.~Fei-Fei.
\newblock {ImageNet:} a large-scale hierarchical image database.
\newblock In {\em Proceedings of the IEEE Conference on Computer Vision and
  Pattern Recognition (CVPR)}, pages 248--255, 2009.

\bibitem{Pascalvoc}
M.~Everingham, L.~Van~Gool, C.~K. Williams, J.~Winn, and A.~Zisserman.
\newblock {The Pascal Visual Object Classes (VOC) Challenge}.
\newblock {\em International Journal of Computer Vision (IJCV)},
  88(2):303--338, 2010.

\bibitem{HariharanICCV2011}
B.~Hariharan, P.~Arbel{\'a}ez, L.~Bourdev, S.~Maji, and J.~Malik.
\newblock Semantic contours from inverse detectors.
\newblock In {\em Proceedings of the IEEE International Conference on Computer
  Vision (ICCV)}, pages 991--998, 2011.

\bibitem{transfernet}
S.~Hong, J.~Oh, B.~Han, and H.~Lee.
\newblock Learning transferrable knowledge for semantic segmentation with deep
  convolutional neural network.
\newblock In {\em Proceedings of the IEEE Conference on Computer Vision and
  Pattern Recognition (CVPR)}, pages 3204 -- 3212, 2016.

\bibitem{HongCVPR17}
S.~Hong, D.~Yeo, S.~Kwak, H.~Lee, and B.~Han.
\newblock Weakly supervised semantic segmentation using web-crawled videos.
\newblock In {\em Proceedings of the IEEE Conference on Computer Vision and
  Pattern Recognition (CVPR)}, pages 7322--7330, 2017.

\bibitem{KhorevaCVPR17}
A.~Khoreva, R.~Benenson, J.~Hosang, M.~Hein, and B.~Schiele.
\newblock Simple does it: Weakly supervised instance and semantic segmentation.
\newblock In {\em Proceedings of the IEEE Conference on Computer Vision and
  Pattern Recognition (CVPR)}, pages 876--885, 2017.

\bibitem{Adamsolver}
D.~P. Kingma and J.~Ba.
\newblock Adam: {A} method for stochastic optimization.
\newblock In {\em Proceedings of the International Conference on Learning
  Representations (ICLR)}, 2015.

\bibitem{sec}
A.~Kolesnikov and C.~H. Lampert.
\newblock Seed, expand and constrain: Three principles for weakly-supervised
  image segmentation.
\newblock In {\em Proceedings of the European Conference on Computer Vision
  (ECCV)}, pages 695--711, 2016.

\bibitem{denseCRF}
P.~Kr\"{a}henb\"{u}hl and V.~Koltun.
\newblock Efficient inference in fully connected crfs with gaussian edge
  potentials.
\newblock In {\em Proceedings of the Neural Information Processing Systems
  (NIPS)}, pages 109--117. 2011.

\bibitem{Alexnet}
A.~Krizhevsky, I.~Sutskever, and G.~E. Hinton.
\newblock {ImageNet} classification with deep convolutional neural networks.
\newblock In {\em Proceedings of the Neural Information Processing Systems
  (NIPS)}, 2012.

\bibitem{KwakAAAI17}
S.~Kwak, S.~Hong, and B.~Han.
\newblock Weakly supervised semantic segmentation using superpixel pooling
  network.
\newblock In {\em Proceedings of the AAAI Conference on Artificial Intelligence
  (AAAI)}, pages 4111--4117, 2017.

\bibitem{scribblesup}
D.~Lin, J.~Dai, J.~Jia, K.~He, and J.~Sun.
\newblock Scribblesup: Scribble-supervised convolutional networks for semantic
  segmentation.
\newblock In {\em Proceedings of the IEEE Conference on Computer Vision and
  Pattern Recognition (CVPR)}, pages 3159--3167, 2016.

\bibitem{Lin16}
G.~Lin, C.~Shen, A.~{van dan Hengel}, and I.~Reid.
\newblock Efficient piecewise training of deep structured models for semantic
  segmentation.
\newblock In {\em Proceedings of the IEEE Conference on Computer Vision and
  Pattern Recognition (CVPR)}, pages 3194 -- 3203, 2016.

\bibitem{Mscoco}
T.-Y. Lin, M.~Maire, S.~Belongie, J.~Hays, P.~Perona, D.~Ramanan,
  P.~Doll{\'a}r, and C.~L. Zitnick.
\newblock {Microsoft COCO:} common objects in context.
\newblock In {\em Proceedings of the European Conference on Computer Vision
  (ECCV)}, pages 740--755, 2014.

\bibitem{LiuCVPR07}
T.~Liu, J.~Sun, N.~N. Zheng, X.~Tang, and H.~Y. Shum.
\newblock Learning to detect a salient object.
\newblock In {\em Proceedings of the IEEE Conference on Computer Vision and
  Pattern Recognition (CVPR)}, pages 1--8, June 2007.

\bibitem{Fcn}
J.~Long, E.~Shelhamer, and T.~Darrell.
\newblock Fully convolutional networks for semantic segmentation.
\newblock In {\em Proceedings of the IEEE Conference on Computer Vision and
  Pattern Recognition (CVPR)}, pages 3431 -- 3440, 2015.

\bibitem{randomwalk}
L.~Lovász.
\newblock Random walks on graphs: A survey, 1993.

\bibitem{BSDS}
D.~Martin, C.~Fowlkes, D.~Tal, and J.~Malik.
\newblock A database of human segmented natural images and its application to
  evaluating segmentation algorithms and measuring ecological statistics.
\newblock In {\em Proc. 8th Int'l Conf. Computer Vision}, volume~2, pages
  416--423, July 2001.

\bibitem{deconvnet}
H.~Noh, S.~Hong, and B.~Han.
\newblock Learning deconvolution network for semantic segmentation.
\newblock In {\em Proceedings of the IEEE International Conference on Computer
  Vision (ICCV)}, pages 1520 -- 1528, 2015.

\bibitem{OhCVPR17}
S.~J. Oh, R.~Benenson, A.~Khoreva, Z.~Akata, M.~Fritz, and B.~Schiele.
\newblock Exploiting saliency for object segmentation from image level labels.
\newblock In {\em Proceedings of the IEEE Conference on Computer Vision and
  Pattern Recognition (CVPR)}, pages 4410--4419, 2017.

\bibitem{Oquab2014}
M.~Oquab, L.~Bottou, I.~Laptev, and J.~Sivic.
\newblock Learning and transferring mid-level image representations using
  convolutional neural networks.
\newblock In {\em Proceedings of the IEEE Conference on Computer Vision and
  Pattern Recognition (CVPR)}, 2014.

\bibitem{wssl}
G.~Papandreou, L.-C. Chen, K.~Murphy, and A.~L. Yuille.
\newblock Weakly-and semi-supervised learning of a {DCNN} for semantic image
  segmentation.
\newblock In {\em Proceedings of the IEEE International Conference on Computer
  Vision (ICCV)}, pages 1742 -- 1750, 2015.

\bibitem{Ccnn}
D.~Pathak, P.~Kr{\"{a}}henb{\"{u}}hl, and T.~Darrell.
\newblock Constrained convolutional neural networks for weakly supervised
  segmentation.
\newblock In {\em Proceedings of the IEEE International Conference on Computer
  Vision (ICCV)}, pages 1742 -- 1750, 2015.

\bibitem{Wsl}
P.~O. Pinheiro and R.~Collobert.
\newblock From image-level to pixel-level labeling with convolutional networks.
\newblock In {\em Proceedings of the IEEE Conference on Computer Vision and
  Pattern Recognition (CVPR)}, pages 1713 -- 1721, 2015.

\bibitem{youtube-obj}
A.~Prest, C.~Leistner, J.~Civera, C.~Schmid, and V.~Ferrari.
\newblock Learning object class detectors from weakly annotated video.
\newblock In {\em Proceedings of the IEEE Conference on Computer Vision and
  Pattern Recognition (CVPR)}, pages 3282 -- 3289, 2012.

\bibitem{Qi2016}
G.-J. Qi.
\newblock Hierarchically gated deep networks for semantic segmentation.
\newblock In {\em Proceedings of the IEEE Conference on Computer Vision and
  Pattern Recognition (CVPR)}, June 2016.

\bibitem{Vgg16}
K.~Simonyan and A.~Zisserman.
\newblock Very deep convolutional networks for large-scale image recognition.
\newblock In {\em Proceedings of the International Conference on Learning
  Representations (ICLR)}, 2015.

\bibitem{OICR}
P.~Tang, X.~Wang, X.~Bai, and W.~Liu.
\newblock Multiple instance detection network with online instance classifier
  refinement.
\newblock In {\em Proceedings of the IEEE Conference on Computer Vision and
  Pattern Recognition (CVPR)}, pages 3059--3067, July 2017.

\bibitem{Tokmakov16}
P.~Tokmakov, K.~Alahari, and C.~Schmid.
\newblock Weakly-supervised semantic segmentation using motion cues.
\newblock In {\em Proceedings of the European Conference on Computer Vision
  (ECCV)}, pages 388--404, 2016.

\bibitem{Vernaza_2017_CVPR}
P.~Vernaza and M.~Chandraker.
\newblock Learning random-walk label propagation for weakly-supervised semantic
  segmentation.
\newblock In {\em Proceedings of the IEEE Conference on Computer Vision and
  Pattern Recognition (CVPR)}, July 2017.

\bibitem{WeiCVPR17}
Y.~Wei, J.~Feng, X.~Liang, M.-M. Cheng, Y.~Zhao, and S.~Yan.
\newblock Object region mining with adversarial erasing: A simple
  classification to semantic segmentation approach.
\newblock In {\em Proceedings of the IEEE Conference on Computer Vision and
  Pattern Recognition (CVPR)}, 2017.

\bibitem{ZifengArxiv16}
Z.~Wu, C.~Shen, and A.~van~den Hengel.
\newblock Wider or deeper: Revisiting the resnet model for visual recognition.
\newblock {\em arXiv preprint arXiv:1611.10080}, 2016.

\bibitem{Crfrnn}
S.~Zheng, S.~Jayasumana, B.~Romera-Paredes, V.~Vineet, Z.~Su, D.~Du, C.~Huang,
  and P.~Torr.
\newblock Conditional random fields as recurrent neural networks.
\newblock In {\em Proceedings of the IEEE International Conference on Computer
  Vision (ICCV)}, pages 1529 -- 1537, 2015.

\bibitem{Cam}
B.~Zhou, A.~Khosla, A.~Lapedriza, A.~Oliva, and A.~Torralba.
\newblock Learning deep features for discriminative localization.
\newblock In {\em Proceedings of the IEEE Conference on Computer Vision and
  Pattern Recognition (CVPR)}, pages 2921 -- 2929, 2016.

\end{thebibliography}

}
\clearpage

%%% APPENDIX %%%%%%%%%%%%%%%%%%%%%%%%%%%%%%%%%%%%%%%%%%
% !TEX root = weaksup_jiwoon.tex

\appendix
\section{Appendix}
\label{sec:appendix}

This appendix provides contents omitted in the regular sections for the sake of brevity.
Section~\ref{app_architecture} and~\ref{app_inference} present technical details of the proposed framework.
Also, in-depth analyses on the quantitative results of our framework are provided in Section~\ref{app_hyperparam} and~\ref{app_fullsup}.
Finally, more qualitative results are given in Section~\ref{app_qualitative}.
We conclude the appendix with brief remarks about future work.

\subsection{Architecture Details of Our Networks}
\label{app_architecture}

\begin{figure}
\centering
\vspace{1.5cm}
\includegraphics[width = \linewidth]{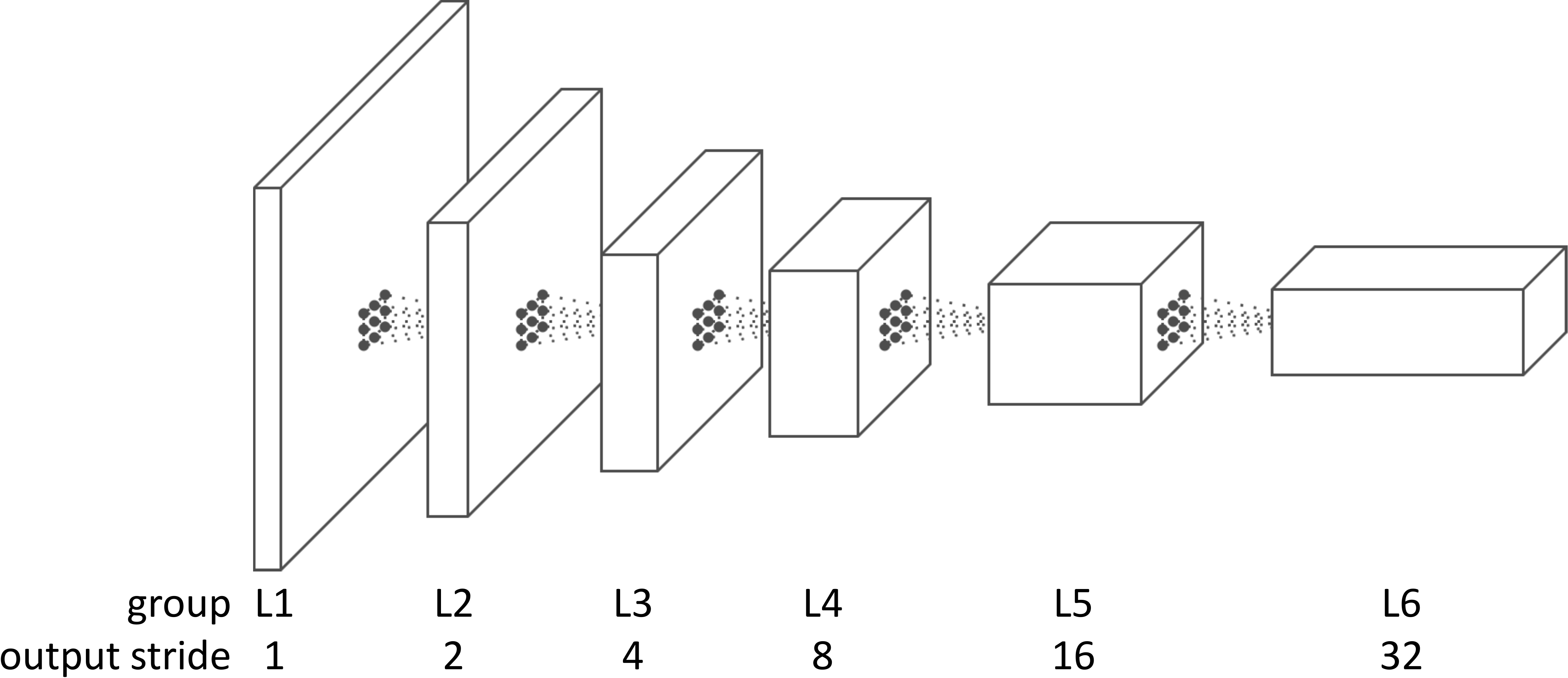}\\
\vspace{0.1cm}
\small{(a) ResNet38}\\
\vspace{0.4cm}
\includegraphics[width = \linewidth]{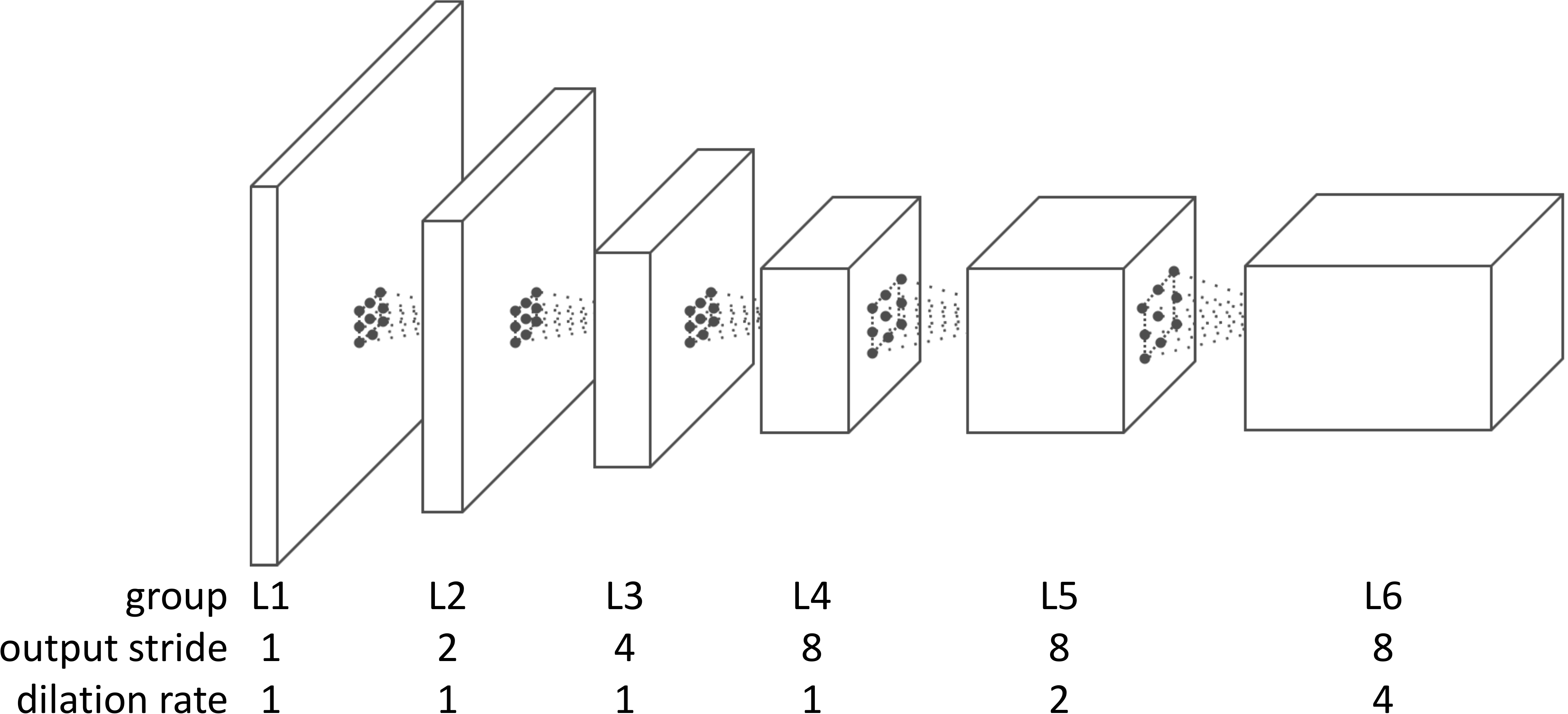}\\
\vspace{0.1cm}
\small{(b) Our backbone network}\\
\vspace{0.4cm}
\includegraphics[width = 0.9 \linewidth]{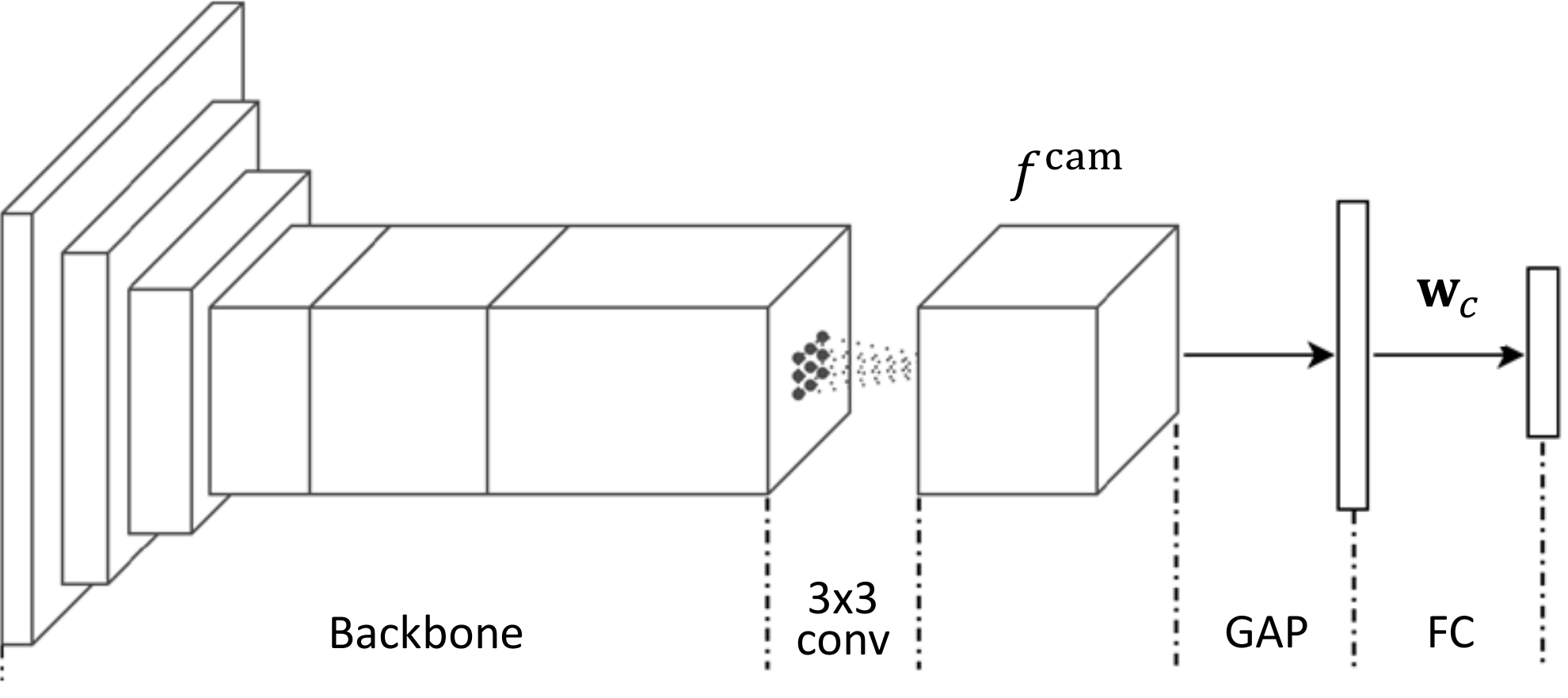}\\
\vspace{0.1cm}
\small{(c) Our network for computing CAMs}\\
\vspace{0.4cm}
\includegraphics[width = \linewidth]{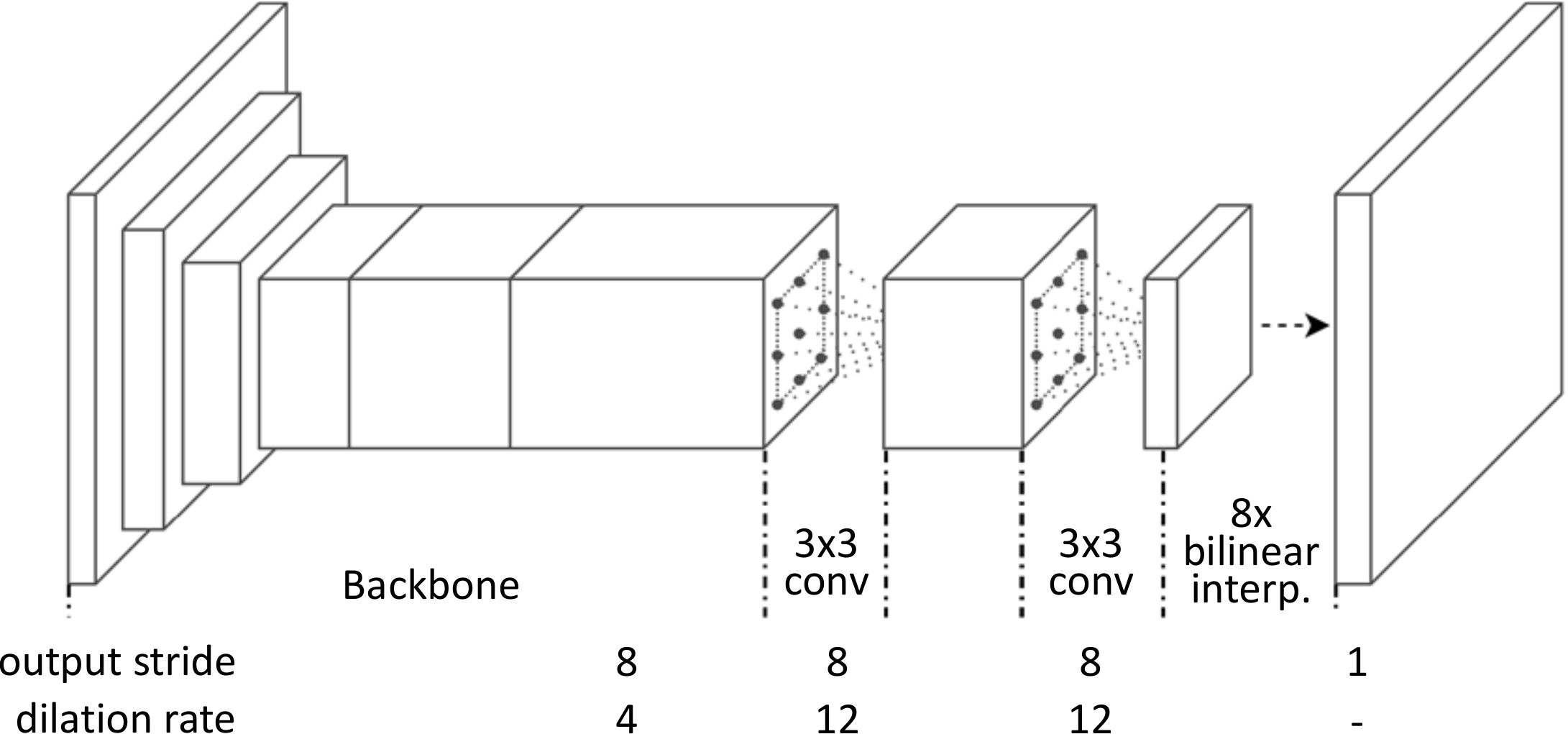}\\
\vspace{0.1cm}
\small{(d) Our semantic segmentation network}\\
\vspace{0.4cm}
\caption{Illustration of detailed network architectures.}
% \vspace{0.7cm}
\label{fig:network_archs}
\end{figure}

This section describes architecture details of the networks in our framework. First, the backbone network is based on ResNet38 proposed in~\cite{ZifengArxiv16} (Figure~\ref{fig:network_archs}(a)), and as illustrated in Figure~\ref{fig:network_archs}(b), its last three levels of convolution layers (L4, L5, and L6) are converted to atrous convolutions. Note that by doubling and quadrupling the dilation rates of L5 and L6, respectively, the output stride of the last convolution feature map of the backbone becomes 8, which is 4 times smaller than that of ResNet38 and enhances the performance of our networks relying on the feature map.

Figure~\ref{fig:network_archs}(c) visualizes the architecture of the network computing CAMs, which is obtained by adding the following three layers on the top of the backbone network: A 3x3 convolution layer for adaptation, a global average pooling (GAP) layer for abstracting the last feature map into a single vector, and a fully connected (FC) layer for learning the entire network with a single classification criterion. Note that the notations in Figure~\ref{fig:network_archs}(c) are set identical to those in Section~\ref{sec:cam} for a clear understanding. Also, the segmentation model of our approach is illustrated in Figure~\ref{fig:network_archs}(d). To obtain this model, we add on the top of the backbone two 3x3 atrous convolution layers with dilation rates of 12, which enable to keep the resolution of the feature maps. In the last stage, a bilinear interpolation is applied to enlarge the resolution of the activation map up to that of the input image. For the architecture of AffinityNet, please refer to Figure~\ref{fig:affinitynet}.

\subsection{Practical Details for Inference}
\label{app_inference}

In this section, we introduce some practical details used to enhance the performance of our segmentation model. Since the segmentation model is trained with randomly flipped and scaled images, we applied the same jittering techniques to images during testing too. Specifically, each test image is horizontally flipped, and rescaled by 5 predefined ratios: 1/2, 3/4, 1, 5/4, and 3/2. As a result, we obtain 10 different versions of a test image, and feed them to the segmentation model to obtain 10 score maps per class accordingly. The score maps are then aggregated by pixel-wise average-pooling to obtain a single map per class, and the final segmentation output is obtained by selecting the class label associated with the maximum score at every pixel. This technique improved the segmentation performance of our model slightly (less than 2\% in mIoU) on the PASCAL VOC 2012 benchmark \emph{val} set.

\iffalse
%% ======================================================================
%% FIGURE START
\begin{figure*}[!htb]
\centering
\small{
\includegraphics[width = 0.495 \textwidth]{figures/fig_backbone_org.pdf}
\includegraphics[width = 0.495 \textwidth]{figures/fig_backbone_dilated.pdf}\\
(a) ResNet38\hspace{6cm}(b) Our backbone network
\includegraphics[width = 0.475 \textwidth]{figures/fig_cam_net.pdf}
\hspace{0.02 \textwidth}
\includegraphics[width = 0.495 \textwidth]{figures/fig_seg_net3.pdf}\\
(c) Our network for computing CAMs\hspace{3cm}(d) Our semantic segmentation network
}
\caption{Illustration of detailed network architectures} 
\vspace{-0.5cm}
\label{fig:architectures}
\end{figure*}
%% FIGURE END
%% ======================================================================
\fi

%% ======================================================================
%% FIGURE START
%% FIGURE END
%% =========================================================

%% ======================================================================
%% FIGURE START
\begin{figure*}%[!ht]
\centering
\includegraphics[width = 0.86 \linewidth]{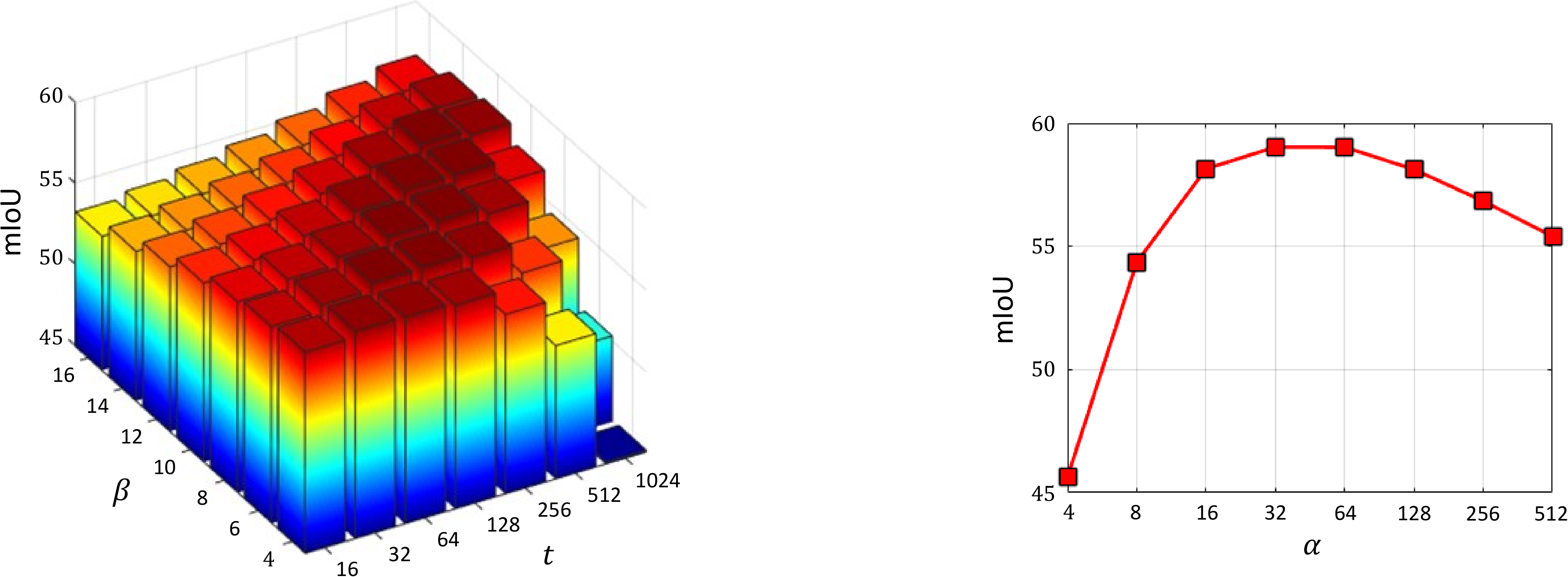}
\hspace{0.14 \linewidth}\\
\vspace{0.15cm}
\hspace{0.17cm}
\small{(a) Accuracy versus $\beta$ and $t$}
\hspace{6cm}
\small{(b) Accuracy versus $\alpha$}\\
\vspace{0.5cm}
\caption{Accuracy (mIoU) of segmentation labels synthesized by CAM+RW for different hyper-parameter values on the VOC 2012 \textit{train}.}
\label{fig:hyperparams}
\end{figure*}
%% FIGURE END
%% ======================================================================

%% ======================================================================
%% TABLE START

\begin{table}[!ht] \small
\centering
\begin{tabular}{ccc|c}
\hline
CAM & CAM+RW & CAM+RW+dCRF & Ours-ResNet38 \\
\hline
46.8 & 57.0 & 58.7 & 61.7 \\ %46.816,57.045,58.670
\hline
\end{tabular}
\vspace{0.4cm}
\caption{Accuracy  of  synthesized  segmentation  labels and our final model in  mIoU, evaluated on the VOC 2012 \emph{val} set.}
\label{tab:comparison_synth}
\vspace{-0.1cm}
\end{table}

%% TABLE END
%% ======================================================================

\subsection{Analysis on Effects of the Hyper-parameters}
\label{app_hyperparam}

We analyzed effects of the hyper-parameters on the quality of synthetic segmentation labels, and the results summarized in Figure~\ref{fig:hyperparams} demonstrate that our label synthesis method is fairly insensitive to the hyper-parameters. 
As shown in Figure~\ref{fig:hyperparams}(a), the quality of synthetic labels fluctuates within only 1.0 mIoU 
% when $\beta$ and $t$ satisfy $\beta \approx 2\log t - 8$.
when $\beta \approx 2\log t - 8$.
In Figure~\ref{fig:hyperparams}(b), we found that the accuracy is saturated when $\alpha$ is greater than 16.
Thanks to this robustness, we believe that little effort will be required to tune the hyper-parameters for other datasets.
% like MS-COCO.

Note that the parameter values given in the paper are not optimal. 
For example, $\alpha$ was 16 in the paper, but it turns out that 32 is better (Figure~\ref{fig:hyperparams}(b)).
This happens because hyper-parameter analyses like Figure~\ref{fig:hyperparams} are prohibited in our setting where no ground-truth segmentation label is given.
Thus for tuning $\alpha$, $\beta$, and $t$, we sampled a small subset of training images and evaluated their effects qualitatively on the subset.
For the same reason, we used the default values given in the original code for the dCRF parameters.

% \subsection{Accuracy of the Synthesized Labels on the Validation Set}
\subsection{Justification of Learning the Fully Supervised Segmentation Network}
\label{app_fullsup}

To empirically demonstrate the advantage of learning a fully supervised segmentation network with synthetic segmentation labels, we compare the accuracy of synthetic segmentation labels and that of our final model on the VOC 2012 \emph{val} set. 

As shown in Table~\ref{tab:comparison_synth}, our synthetic segmentation labels, denoted by CAM+RW+dCRF, achieves 58.7 mIoU while the final segmentation network attains 61.7.
The gap between them justifies our strategy of learning a segmentation model.
Note that, in the above comparison, CAM+RW+dCRF gets an unfair advantage over the segmentation model since it utilizes ground-truth image-level labels to filter out CAMs of irrelevant classes (Section~\ref{sec:cam}).
Without such labels, the score of CAM+RW+dCRF will be even further degraded.

\subsection{More Qualitative Results of Our Approach}
\label{app_qualitative}
\begin{figure*}[!ht]
\small
 \hspace{0.7cm}Input Image\hspace{1.2cm} Ground-truth \hspace{1.5cm} CAM \hspace{1.3cm} Semantic Affinity \hspace{1.0cm} CAM+RW \hspace{0.9cm} CAM+RW+dCRF
\vspace{-0.18cm}
\begin{center}
\includegraphics[width=1 \linewidth] {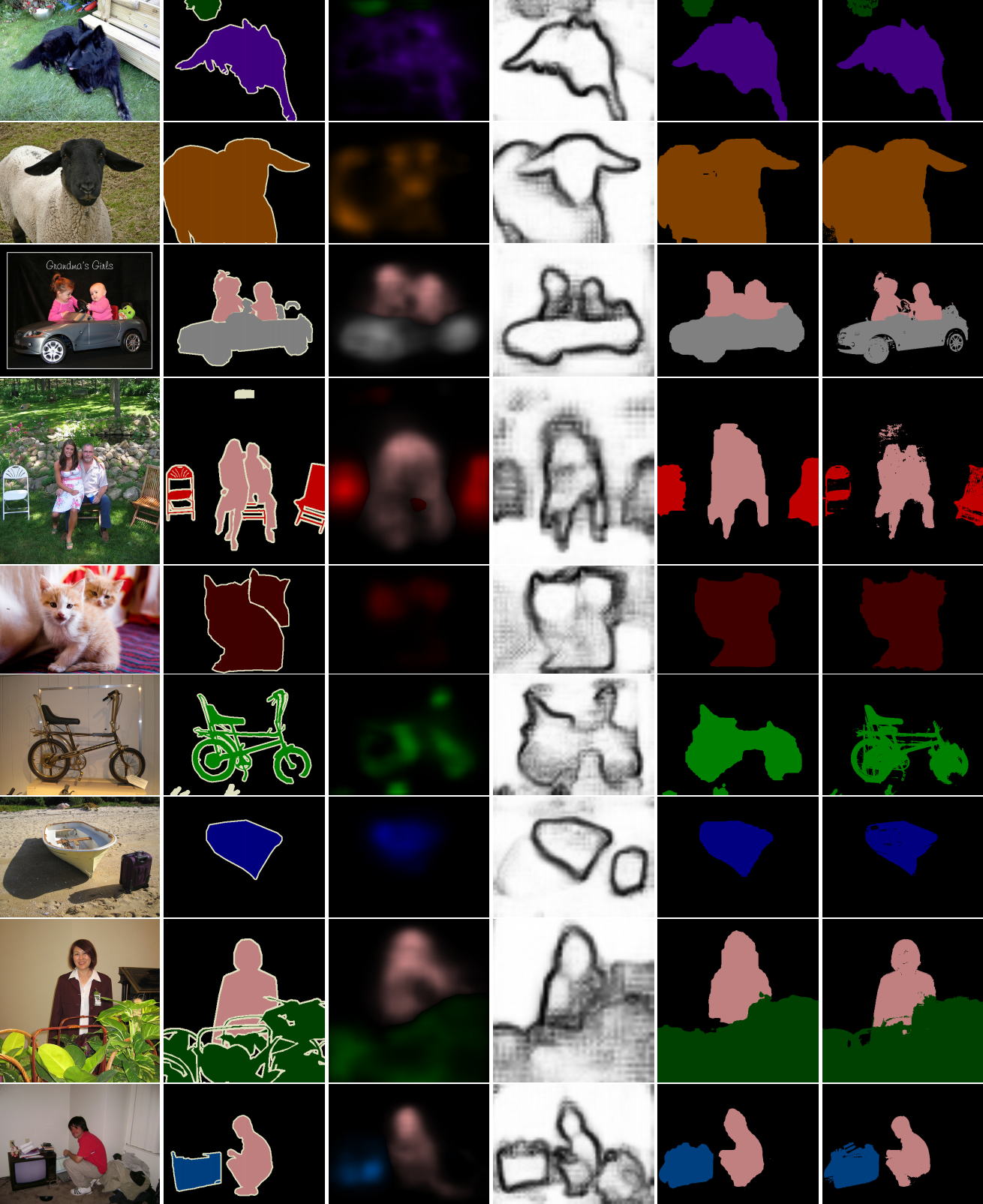}
\end{center}
\caption{Qualitative Examples of Synthesized Segmentation Labels on the PASCAL VOC 2012 \emph{train} images.}
\label{fig:app_qual1}
\end{figure*}

\begin{figure*}[!ht]
\small
 \hspace{0.7cm}Input Image\hspace{1.2cm} Ground-truth\hspace{1.0cm} Ours-ResNet38 \hspace{1.0cm} Input Image \hspace{1.2cm} Ground-truth\hspace{1.0cm} Ours-ResNet38
\vspace{-0.18cm}
\begin{center}
\includegraphics[width=1 \linewidth] {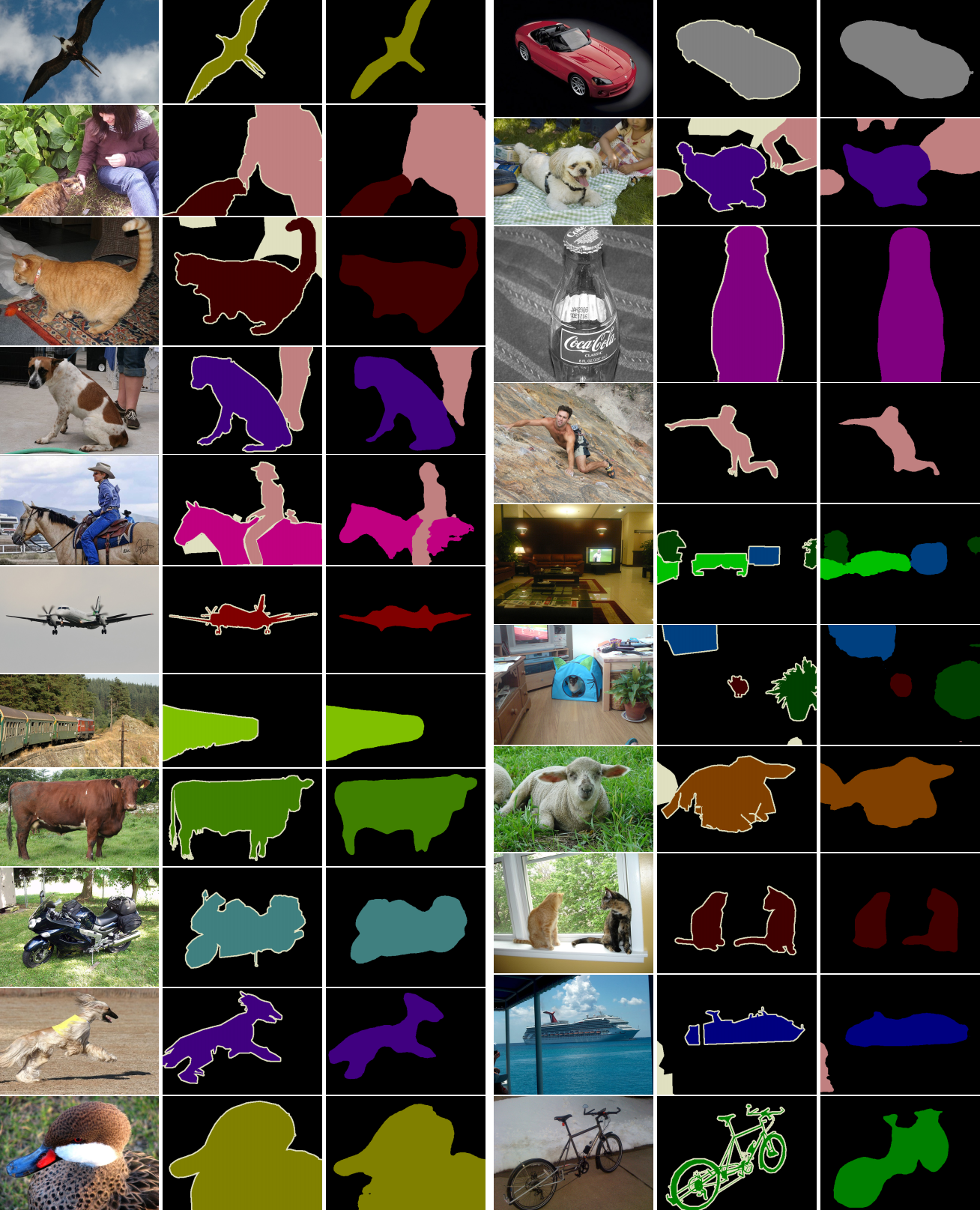}
\end{center}
\caption{Semantic Segmentation Results on the PASCAL VOC 2012 \emph{val} images. }
\label{fig:app_qual2}
\end{figure*}

We provide more qualitative results omitted in the regular sections for the space limit. 
Figure~\ref{fig:app_qual1} illustrates our segmentation label synthesis procedure with a number of qualitative examples. 
More segmentation results of our final model (\ie, Ours-ResNet38) are presented in Figure~\ref{fig:app_qual2}.

\subsection{Future Work}
Next on our agenda is to utilize AffinityNet in a transfer learning setting where source domain provides groundtruth segmentation labels for exclusive object classes.
By leveraging such external data for training AffinityNet, our framework could generate more accurate segmentation labels in target domain as AffinityNet learns and predicts class-agnotic affinities that can be generally applicable.
Another direction for further exploration is weakly supervised semantic boundary detection using AffinityNet, which already shows its potential in the experimental results.

\end{document}